\documentclass[lettersize,journal]{IEEEtran}
\usepackage{amsmath,amsfonts}
\usepackage{algorithmic}
\usepackage{algorithm}
\usepackage{array}
\usepackage{textcomp}
\usepackage{stfloats}
\usepackage{url}
\usepackage{verbatim}
\usepackage{graphicx}
\usepackage{cite}
\usepackage{bm}
\usepackage{gensymb}
\usepackage{subfigure}
\usepackage{booktabs}
\usepackage{multirow}
\usepackage{color}
\usepackage{threeparttable}
\usepackage{hyperref}
\usepackage{dsfont}
\usepackage{tabularray}
\usepackage{caption}
\begin{document}

\title{YOPOv2-Tracker: An End-to-End Agile Tracking and Navigation Framework from Perception to Action}

\author{Junjie Lu, Yulin Hui, Xuewei Zhang, Wencan Feng, Hongming Shen, Zhiyu Li, and Bailing Tian*}

\twocolumn[{
\renewcommand\twocolumn[1][]{#1}
\maketitle
\begin{center}
    \captionsetup{type=figure}
    \vspace{-0.2cm}
    \includegraphics[width=\textwidth]{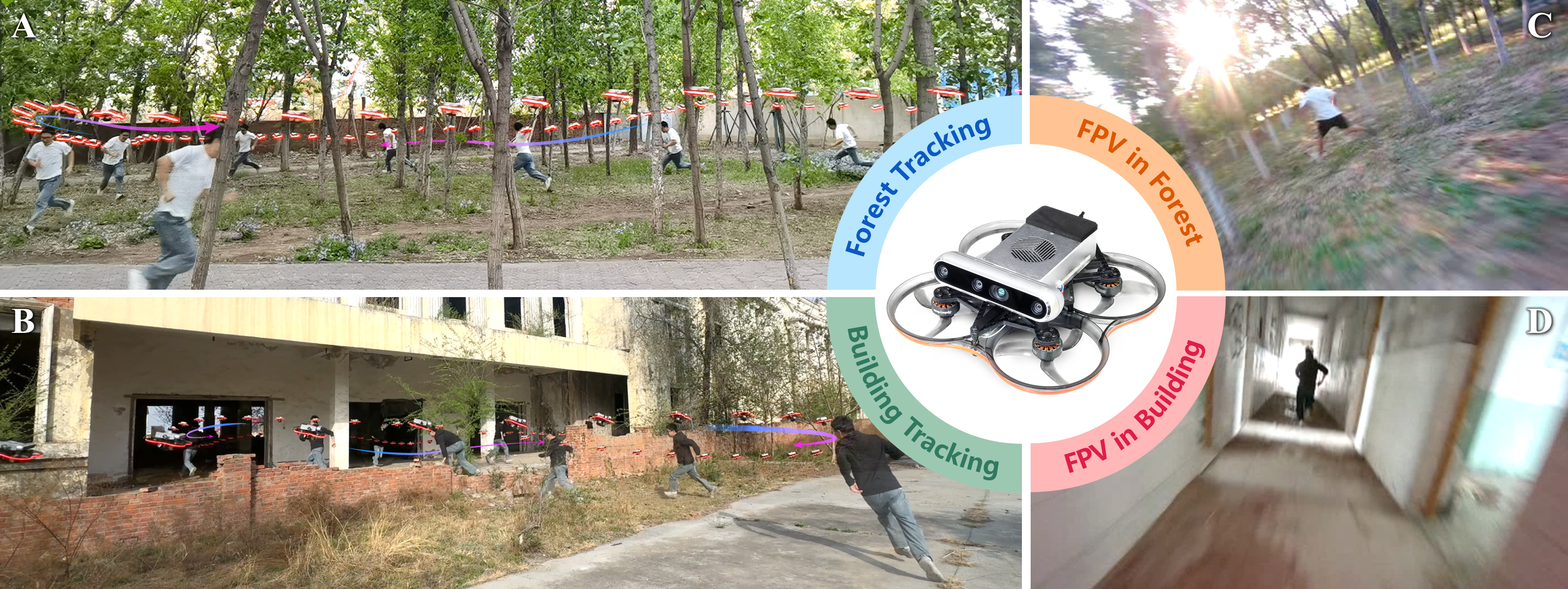}
    \captionof{figure}{Overview of the proposed aerial tracking and navigation system. (A-B) show the agile tracking in dense forests and cluttered buildings, while (C-D) depict the first-person view (FPV) of the tracker. The system is fully autonomous and onboard, with further details provided in Section \ref{Sec_Experiment} and attached video.}
    \label{cover}
\end{center}
}]

\let\thefootnote\relax\footnotetext{All authors are with the School of Electrical and Information Egineering, Tianjin University, Tianjin, 300072, China.}

\begin{abstract}
To achieve efficient tracking of unpredictable target in cluttered environments, a series of improvements in detection, mapping, navigation, and control are introduced in previous work to make the overall system more comprehensive. However, this separated pipeline introduces significant latency and limits the agility of quadrotors, particularly on computationally constrained onboard devices. On the contrary, we follow the design principle of ``less is more'', striving to simplify the process while maintaining effectiveness. In this work, we propose an end-to-end agile tracking and navigation framework for quadrotors that directly maps the sensory observations to control commands. Importantly, leveraging the multimodal nature of navigation and detection tasks, our network maintains interpretability by explicitly integrating the independent modules of the traditional pipeline, rather than a crude action regression. In detail, we adopt a set of motion primitives as anchors to cover the searching space regarding the feasible region and potential target. Then we reformulate the trajectory optimization as regression of primitive offsets and associated costs considering the safety, smoothness, and other metrics. For tracking task, the trajectories are expected to approach the target and additional objectness scores are predicted. Subsequently, the predictions, after compensation for the estimated lumped disturbance, are transformed into thrust and attitude as control commands for swift response. During training, we seamlessly integrate traditional motion planning with deep learning by directly back-propagating the gradients of trajectory costs to the network, eliminating the need for expert demonstration in imitation learning and providing more direct guidance than reinforcement learning. Finally, we deploy the algorithm on a compact quadrotor and conduct real-world validations in both forest and building environments to demonstrate the efficiency of the proposed method. For supplementary video see {\href{https://youtu.be/QBWEDoQ4xaQ}{https://youtu.be/QBWEDoQ4xaQ}}. The code will be released at \href{https://github.com/TJU-Aerial-Robotics/YOPO-Tracker}{https://github.com/TJU-Aerial-Robotics/YOPO-Tracker}.
\end{abstract}


\section{Introduction}
\IEEEPARstart{W}{ith} the development of autonomous navigation and deep learning, unmanned aerial vehicles (UAVs) show great potential in various applications and have been widely used in more and more complex missions. Agile tracking of moving target (as shown in Fig. \ref{cover}) is a promising field to promote the applications of autonomous quadrotors in tasks such as aerial photography, pursuit, human-robot interaction, and air-ground coordination. A typical solution is to decompose the tracking problem into detection, mapping, planning, and control as independent subtasks, enabling the adoption of well-established methods for each module and making the overall system more interpretable. However, the sequential approach introduces significant latency and compounding errors, making it fatal for high-speed flight and agile tracking in cluttered environments, particularly for compact quadrotors with limited visual sensors and onboard computational resources.

In this work, we concentrate on achieving autonomous obstacle avoidance and tracking of fast-moving target by UAV in obstacle-dense environments, relying only on limited visual sensors and onboard computational resources. The current research on target tracking \cite{elastic-tracker,visibility-tracker,perching-tracking} primarily addresses the problem of trajectory generation to follow the target, while treating detection and control as off-the-shelf modules or even using the ground-truth as priors. Furthermore, the trajectory generation problem is usually separated into (i) perception and mapping, (ii) front-end path searching, and (iii) back-end trajectory optimization. Firstly, an efficient perception and mapping module is essential to provide safety constraints and filter sensor noise by maintaining a continuously updated occupancy grid map. Afterwards, a searching method that considers both target observation and obstacle avoidance is usually adopted as the front end to provide a rough initial path. Finally, the initial path is further improved by the back-end optimization considering the safety, feasibility, and target visibility. Although such separated pipelines have achieved impressive results in autonomous navigation, their sequential nature introduces additional latency, making it challenging for high-speed and agile maneuvers. In contrast, recent learning-based approaches \cite{science, Newton, yopo} achieve superior performance by tightly integrating robust perception with efficient planning. By predicting navigation commands directly from sensor measurements, they significantly decrease the latency between perception and action, and indicate that the intermediate stages such as mapping and planning are not essential. However, existing end-to-end policies primarily focus on pushing the speed limits of quadrotors in racing or navigation, whereas more complex missions such as target tracking remain open research problems in aerial robotics. 

\begin{figure}[t]\centering
        \includegraphics[width=\linewidth]{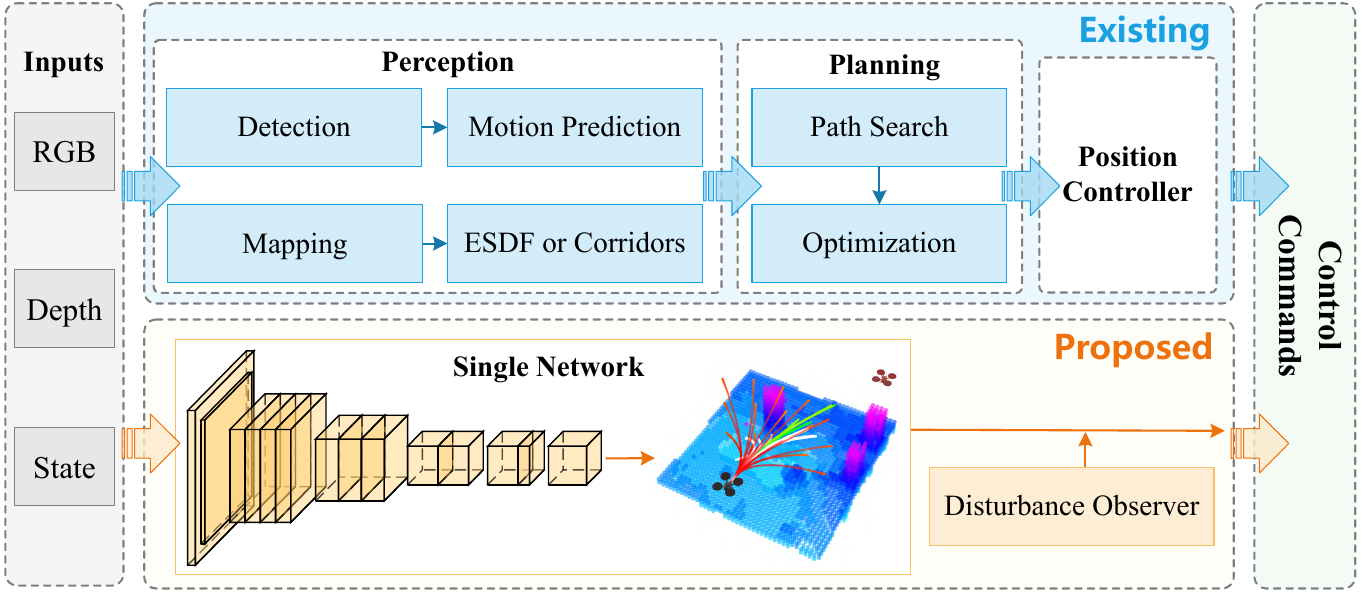}
        \caption{Comparison with existing tracking pipeline.}\label{introduction}
\end{figure}

In this work, we propose a one-stage end-to-end tracker YOPOv2-Tracker (You Only Plan Once), which further integrates object detection into the navigation policy by leveraging the nature that the solutions of navigation and detection tasks are both multi-modal. That is, the target may appear at arbitrary locations in the image, and multiple feasible trajectories may exist for obstacle avoidance within the quadrotor's field of view (FOV). To solve this problem, object detection methods \cite{faster-rcnn} typically place anchor boxes at each sliding-window location and use region proposal algorithms to hypothesize potential object locations. Subsequently, the features of each proposal are passed through fully connected layers for classification and bounding-box refinement. Similarly, for navigation tasks \cite{Topological, Tgk-planner}, multiple feasible paths with different topologies are searched as initial values for back-end optimization to avoid getting stuck in suboptimal local minima. After that, the optimization-based methods which incorporate more constraints are applied as the back end for further improvement. Based on this, we extend the anchor boxes in detection to prior primitive trajectories within the FOV of camera to thoroughly explore the feasible space and cover the potential target. Then, a simple fully convolutional network, inspired by the state-of-the-art one-stage detector \cite{yolo}, is developed to predict the offsets and scores of primitives for refinement. Different from our previous work \cite{yopo}, not only the safety and smoothness, but also the tracking performance and objectness scores of predictions are taken into account in the training process. Finally, the non-maximum suppression (NMS), with consideration of the continuity of target estimation, is deployed to filter the predictions which are infeasible or invalid for tracking. 
Moreover, target motion prediction, either fitted using past detections or predicted by an individual network, is typically included in traditional trackers \cite{elastic-tracker, visibility-tracker} and autonomous driving \cite{driving}. However, in practice, targets such as humans are agile and may take deceptive and evasive actions to avoid being tracked. They neither follow a predefined planning strategy nor cooperate like vehicles in autonomous driving. 
As visualized in Fig. \ref{introduction}, instead of making improvements to existing framework to make it more comprehensive and complex, we follow the opposite philosophy of ``less is more" and strive to achieve immediate response through a minimalist architecture, enabling high-speed flight and aggressive tracking in cluttered environments. The network is designed to be simple, straightforward, yet meaningful, rather than a black-box regression of control commands. It explicitly incorporates perception, detection, path search, and optimization of the traditional pipeline within a single forward propagation. To the best of our knowledge, this is the first end-to-end tracker that achieves high-speed tracking of unpredictable target in cluttered real-world environments. 

Leveraging the differential flatness property of quadrotors, most navigation approaches \cite{science, yopo, deep_panther} optimize the polynomial or B-spline trajectory, and discretize it into reference state (the position and its derivatives) at fixed intervals for tracking by the position controller. To ensure the continuity of the reference states in control, the planners take the previous reference state as initial value rather than the current actual state. This is reasonable in mapping-and-planning approaches, as the environmental information at the reference position is contained in map. However, for the end-to-end policies, it can be fatal with the tracking error between reference and actual states increasing in high-speed flight, because only the image at the actual position is observable.
To avoid this inconsistency and decrease the error accumulation in hierarchical control, we further remove the position controller and plan the attitude directly, thereby eliminating the continuity constraint on reference position. To cope with model uncertainty and minimize the sim-to-real gap, domain randomization or precise model identification are commonly leveraged in previous learning-based controllers \cite{Newton, Racing_SR, geles2024demonstrating}. However, such techniques introduce additional challenges, such as the need to apply disturbances or model changes during training, and make real-world deployment more complex. Different from previous methods that predict the control commands directly, we optimize trajectory in differentially flat space with an ideal point-mass physics model and employ disturbance observer to estimate model uncertainty and external disturbances. Subsequently, the lumped disturbance is added to the desired acceleration (i.e., the second derivative of trajectory) with the same dimensionality and converted into the desired attitude. Compared to previous solutions, we simplify the real-world deployment as the policy is independent of the actual physical model and inputs of disturbance observer can be directly obtained from the state estimator. In addition, the proposed pipeline enables the network to focus on exploring patterns between observations and actions, while decoupling uncertain disturbances and the analytical kinodynamic model from the policy. 

To train the network policies, imitation learning and reinforcement learning are utilized by either imitating a privileged expert or exploring the optimal policy through trial-and-error. The imitation-learning-based methods \cite{science, deep_panther, transformer_kumar} have achieved impressive performance in swift flight by approximating a computationally expensive algorithm with a lightweight network. However, due to the multimodal nature of the navigation problem, the distance to the expert demonstrations cannot represent the realistic performance of the predictions, even with the multi-hypothesis winner-takes-all (WTA) loss. Moreover, using explicit labels lacks exploration of the action space and makes the policy heavily dependent on the quality and comprehensiveness of expert demonstrations. 
On the contrary, policies in reinforcement learning \cite{Racing_SR, RL-based2} are trained by maximizing the rewards from the environment through trial-and-error, which realistically reflects the performance of the action and enables better exploration. However, compared to direct supervision, the reward signal of reinforcement learning is frequently sparse, noisy, and delayed, making it struggle with slow convergence and tend to be data-intensive.
To overcome the limitations of previous methodologies, we propose an end-to-end training strategy without any expert demonstrations or simulator interactions. We replace the action-value function, which is typically approximated by the critic network in reinforcement learning, with trajectory cost evaluated by differentiable privileged map for more realistic, accurate, and timely feedback. Extending traditional gradient-based planners that optimize trajectory parameters, we further back-propagate the gradients to the weights of neural network via the chain rule. This expert-free strategy simplifies the training process significantly and makes it possible to predict more candidate trajectories across sliding windows without complex expert demonstrations or label assignments. Additionally, it allows more flexible data augmentation while not requiring re-annotation or re-interaction with the simulator. Leveraging the guidance provided by the pre-built privileged map for direct gradient descent, it improves the training efficiency and is scalable to more complex tasks compared to stochastically sampling the policy in reinforcement learning.

Finally, we deploy the proposed method on a compact 155 mm quadrotor equipped with RealSense D455 for RGB-D perception and NVIDIA Orin NX as onboard computer. We demonstrate superior computational efficiency and agile tracking performance in both cluttered forest and urban environments. During validations, the quadrotor not only needs to handle the evasive maneuvers of the evader but also avoid sudden obstacles, while only limited visual sensor  and onboard computational resources are available. Additionally, by modifying the costs, our framework can be easily extended to other multimodal tasks such as high-speed navigation and obstacle avoidance. 

In summary, the proposed method enables an instinctive, biologically inspired perception-to-action process for agile tracking, while preserving the interpretability of the traditional pipeline. Moreover, it seamlessly bridges classical trajectory optimization and deep learning through end-to-end gradient back-propagation. The main contributions of this work are as follows:
\begin{enumerate}
\item An end-to-end framework for agile tracking and high-speed navigation is proposed, which directly maps the sensory observations to low-level control commands. Leveraging the multimodal nature of obstacle avoidance and detection tasks, we integrate perception, detection, and navigation into a unified network and decrease the processing latency significantly.
\item To reduce error accumulation and simplify the real-world deployment, we directly calculate the desired attitude from the network's predictions, incorporating the model uncertainties and disturbances estimated by the disturbance observer.
\item The network is trained using the numerical gradients directly back-propagated from the privileged environment and deployed onto a compact physical platform for real-world validation. Moreover, the source code and hardware platform are released for the reference of the community.
\end{enumerate}

\section{Related Works}

\subsection{Autonomous Navigation}
Autonomous navigation of quadrotors in cluttered environments is an important prerequisite for target tracking and has been extensively investigated. One of the earliest works \cite{richter} ensures the safety of the piecewise polynomial trajectories by iteratively inserting intermediate waypoints to the colliding segments. Subsequently, \cite{corridor2, corridor1} adopt the idea of safe-flying corridor, in which the trajectories are constrained into an obstacle-free space consisting of multiple convex shapes. By contrast, some other methods optimize trajectories by minimizing the cost function considering the smoothness, safety, and dynamic feasibility simultaneously. Specifically, \cite{gao_gradient, Oleynikova_gradient} optimize the waypoints of the piecewise polynomial trajectory with the gradient of Euclidean signed distance field (ESDF) and dynamic constraints. In comparison, utilizing the convex hull property of B-spline, \cite{fast_planner, ego_planner} optimize the control points of the B-spline to ensure the safety and feasibility of the trajectory. To avoid suboptimal solutions of gradient descent optimization, some topological path searching algorithms are adopted in \cite{Topological, Tgk-planner} as the front end to explore the solution space comprehensively and generate multiple topologically distinct initial paths. To address the non-continuous cost and constraints, \cite{mppi_planner} adopt the gradient-free model predictive path integral framework to approximate the optimal trajectory using forward sampling of stochastic diffusion processes. Although the robustness and reliability of the above methods have been extensively validated and widely applied in both academia and industry, the cascade structure introduces latency and error accumulation, limiting their scalability to high-speed flight. To push the limits of the flight speed, \cite{Bubble} enlarges the corridor spaces of overlapping spheres with novel designs and hence allows the quadrotor to maneuver at higher speeds. Moreover, \cite{ren2025safety} plans trajectories using LiDAR point clouds and extends the planning horizon to invisible spaces. They achieve remarkable performance in autonomous navigation and greatly advance the high-speed flight capabilities of UAVs. However, these achievements depend on high-precision and long-range LiDAR sensors, limiting their applicability for compact UAVs equipped with noisy and constrained visual sensors.

In contrast to these approaches, the breakthroughs of deep learning inspire alternative solutions to autonomous navigation without explicit mapping and planning stages. Benefiting from the data-driven nature of deep learning, \cite{imitating} realizes the earliest learning-based navigation in cluttered natural environments relying only on a monocular camera. Recently, some reinforcement learning-based methods \cite{Racing_SR, geles2024demonstrating} focus on training a controller for drone racing and outperform the human world champion pilots. However, due to the slow convergence and data-intensive nature of reinforcement learning, these applications usually concentrate on specific control tasks without considering perception in previously unknown environments. In contrast, \cite{science} adopts imitation learning to train a lightweight network policy using demonstrations given by a privileged expert, and achieves unprecedented high-speed performance in the wild. Similarly, \cite{deep_panther} improves the label assignments strategy between the expert demonstrations and predictions in imitation learning. In comparison, leveraging state-of-the-art differentiable simulation, \cite{Newton} significantly improves the training efficiency of reinforcement learning and achieves remarkable vision-based swift flight, relying solely on extremely limited airborne resources. Different from predicting the trajectory or control commands, some other methods \cite{lpnet, primitive_net} train a network for collision probability prediction of pre-defined motion primitives. Additionally, some learning-based approaches address problems that are difficult to mathematically model in traditional planning, such as adaptive flight speed in different scenarios to balance the safety and agility \cite{Mavrl, adaptation_speed}, and time allocation for piecewise segments to generate smooth and fast trajectories \cite{time_allo}. In contrast with aforementioned works, we employ a detection-like network to capture the multimodality of navigation and avoid the mode collapse problem caused by directly predicting multiple distinct trajectories with a symmetric architecture. Besides, our model is dynamics-independent, and we address the sim-to-real gap of model uncertainty in end-to-end control through the disturbance observer.

\subsection{Target Tracking}
Compared to obstacle avoidance, the target tracking task introduces additional detection module and tracking constraints for motion planning. In early works \cite{Visual_Servo, Visual_Servo2}, visual servoing was applied to cinematography and aerial photography by utilizing visual errors in images for feedback control to maintain the target within the FOV. However, such methods are limited to applications in obstacle-free scenarios, while planning-based approaches, which incorporate tracking constraints into either front-end search or back-end optimization, are more widespread. Specifically, \cite{fast-tracker} proposes a target-informed kinodynamic searching method as the front end and generates corridor based on the heuristically searched tracking path. On the contrary, \cite{chen2016tracking} generates an initial path parallel to the predicted target trajectory to keep a proper tracking distance, and only takes the tracking cost into account during back-end optimization. Furthermore, \cite{videography-tracker, videography-tracker2} not only consider the target visibility in front-end corridor generation, but also optimize the distance between the quadrotor and viewpoints of predicted target during back-end optimization. Moreover, generative functions are defined in optimization problem in \cite{chen2017using} to keep different relative motion patterns between the quadrotor and target. Considering both observation distance and occlusion effect, \cite{elastic-tracker} proposes an occlusion-aware path search method and guarantees visibility through hard constraint constructed by a series of sector-shaped visible regions around the future waypoints of target. Differently, \cite{visibility-tracker} proposes a differentiable target visibility metric that approximates the expected obstacle-free FOV with a sequence of spherical regions. For more reasonable motion predictions, \cite{intention-tracker} designs a risk assessment function and a state observation function to predict the target intention with the human joints detected by specific pose estimation module. In addition to tracking targets like humans, related technologies can also be applied in fields such as adaptive tracking and perching \cite{perching-tracking}, aerial and ground vehicles coordination \cite{zhang2023coni}, and multi-UAVs capture and confrontation \cite{capturing}. 

However, existing systems either use off-the-shelf target detection modules such as \cite{yolo}, or neglect the detection by treating the target position as prior information, which can be localized by the AprilTag \cite{fast-tracker} or sent by the target directly\cite{visibility-tracker}. Moreover, target motion prediction is typically a critical component for tracking constraint construction in trajectory optimization. Some approaches \cite{chen2016tracking, fast-tracker} predict the target's motion by polynomial fitting or Bezier regression using past detections, while others \cite{videography-tracker, intention-tracker} further consider the dynamic feasibility, collision avoidance, or even the intention for more reasonable predictions. However, targets such as humans are agile and may take deceptive and evasive actions to avoid being tracked. Therefore, the motion prediction of the non-cooperative target is unreliable, especially with high-speed evasive maneuvers, which imposes higher requirements on the real-time performance of the tracking system. As a result, traditional frameworks including detection, mapping, planning, and control are limited to low-speed, sparsely obstructed environments. 

Instead of making improvements to existing framework to make the pipeline more comprehensive and complex, such as incorporating feasibility or intention into target motion prediction, we adopt a completely different design philosophy. We strive to achieve rapid response through a minimalist architecture, enabling high-speed flight and aggressive tracking in cluttered environments. Although some methods \cite{learning-tracking} make pioneering attempts at end-to-end tracking, the crude action regression lacks purposeful design of detection and collision avoidance, limiting its application to the laboratory. In comparison, the proposed method explicitly incorporates detection, search, and optimization of the traditional tracking pipeline into a single network. To the best of our knowledge, it is the first end-to-end tracker that achieves high-speed tracking of unpredictable target in cluttered real-world environments.

\begin{figure*}[t]\centering
        \includegraphics[width=\linewidth]{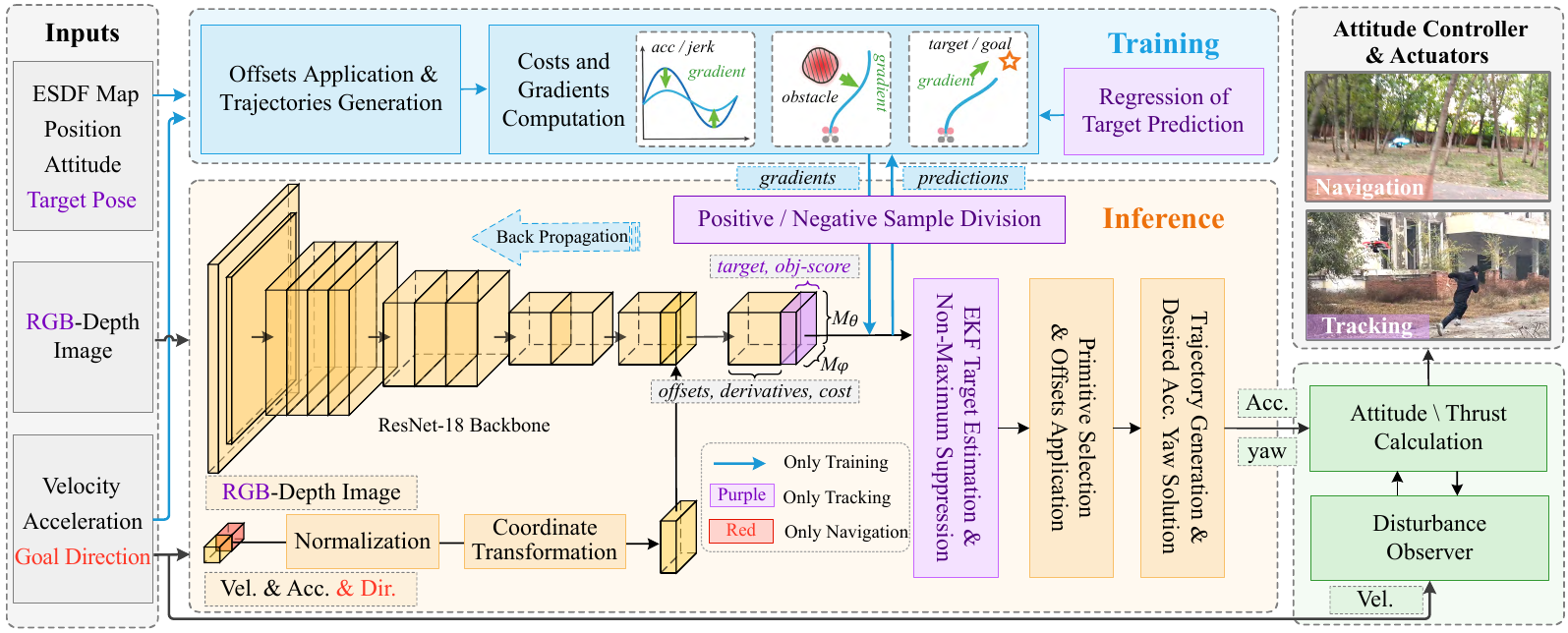}
        \caption{System Overview. The proposed framework takes RGB-D images and state observations as inputs, and outputs attitude and thrust as control commands. The ground-truth map is available in training while only the image observation is accessible during inference. It can easily switch between different tasks by modifying the modules marked in red and purple, which are required only for navigation and tracking, respectively.}\label{system_overview}
        \vspace{-0.2cm}
\end{figure*}

\section{Method}

\subsection{System Overview}
The overall architecture of the proposed navigation and tracking framework is illustrated in Fig. \ref{system_overview}. It takes RGB-D images and state estimates as inputs and outputs attitude and thrust as control commands. The system is designed very simple and straightforward, primarily a lightweight fully convolutional network. Specifically, we use a series of primitives to explore feasible solutions and potential target (Section \ref{Sec_Traj}), and a fully convolutional network to directly predict the safe tracking trajectory (Section \ref{Sec_Network}). Subsequently, the desired control commands are computed while accounting for external disturbances and model uncertainty (Section \ref{Sec_Contorl}). Additionally, we leverage privileged environmental information for prediction evaluation and back-propagate the gradients to the neural network for end-to-end training (Section \ref{Sec_Training}).

\subsection{Trajectory Representation} \label{Sec_Traj}

In contrast to state-of-the-art control-based approaches, we plan trajectories in a reduced space of differentially flat outputs, which inherently guarantees the smoothness and continuity of the flight. Without considering the full dynamics and disturbances, we present the motion as three independent one-segment time-parameterized polynomials and handle yaw separately:
\begin{equation}
	{f_\mu}(t) = {a_0} + {a_1}t + {a_2}{t^2} + {a_3}{t^3} + ... + {a_n}{t^n}\, .
\end{equation}
For brevity, we omit the axis subscript $\mu \in \{x, y, z\}$ in the following. To ensure smooth attitude transformed from acceleration directly (as described in Section \ref{Sec_Contorl}), we use the fifth-order polynomial with second-order continuous initial derivatives $\bm{d}_F = [{f}(0), {\dot f}(0), {\ddot f}(0)]^T$. We fix the execution time $T$ and treat the final derivatives $\bm{d}_P = [{f}(T), {\dot f}(T), {\ddot f}(T)]^T$ as optimized variables. Then, the polynomial can be reformulated in the form of a Hermite curve:
\begin{equation}
        {f}(t) = \bm{tMd} \,.
        \label{trajectory}
\end{equation}
Where $\bm{t} = [1, t, t^2, t^3, t^4, t^5]$, the constant matrix $\bm{M}$ relevant to $T$ is utilized to map the boundary derivatives to the coefficients, and $\bm{d} = [\bm{d}_F, \bm{d}_P]^T$ is the rearranged derivative vector.

\begin{figure}[t]\centering
        \includegraphics[width=0.8\linewidth]{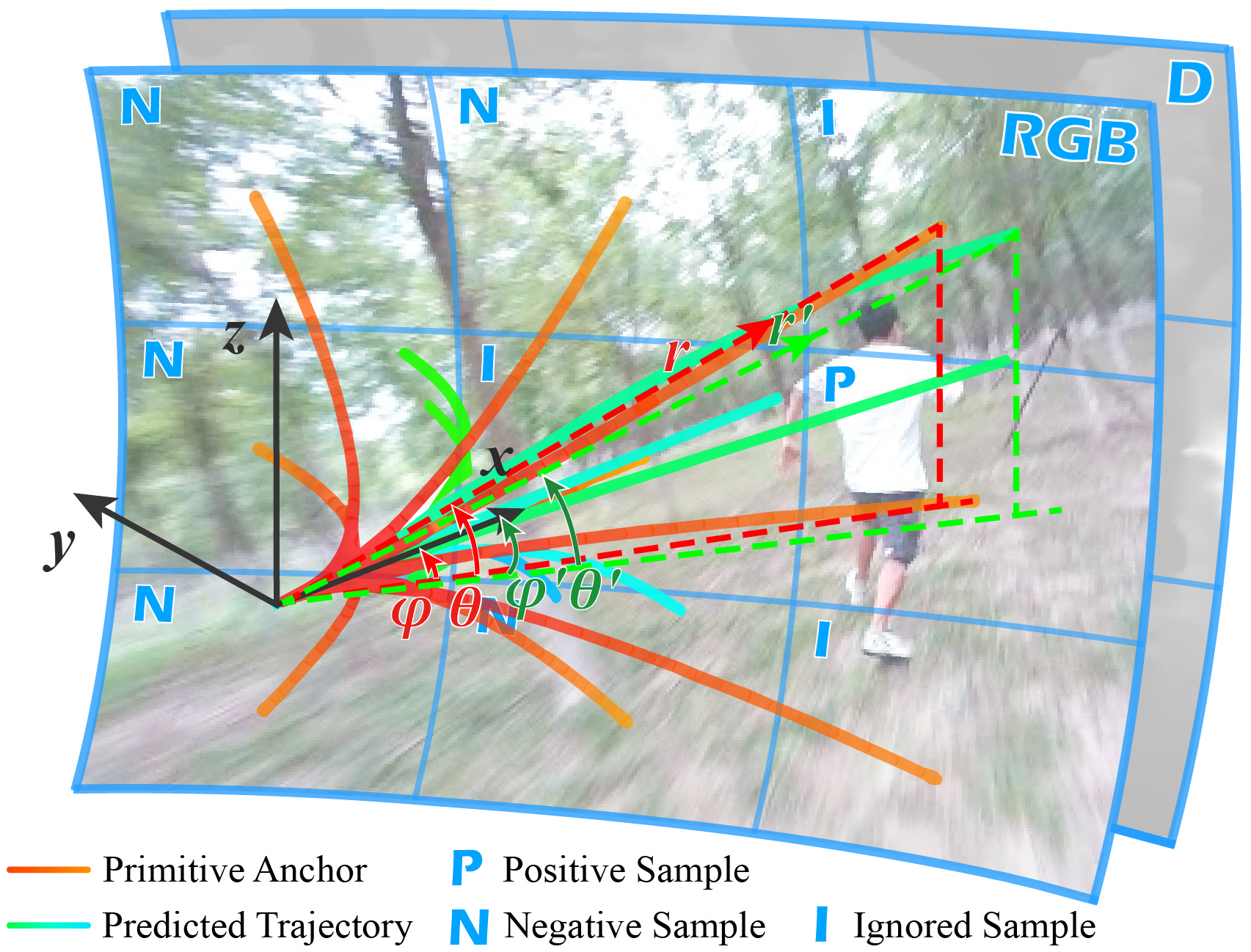}
        \caption{We divide the RGB-D image into grids, each of which corresponds to a primitive and a prediction. In the tracking task, the predictions are further divided into positive, negative, and ignored samples based on the target's location in the image.}\label{symbol}
\end{figure}

To avoid being trapped in suboptimal local minima, the topology-guided front ends \cite{Topological, Tgk-planner} are designed to search the solution space more thoroughly and provide multiple initial paths for local optimization. Similarly, in object detection tasks, a large number of sliding windows and anchor boxes are placed over the image to ensure comprehensive coverage. Given the multimodal characteristic of trajectory planning and object detection, we substitute the prior anchor boxes in detection with primitive-based planners \cite{primitive3, primitive4}. Specifically, as visualized in Fig. \ref{symbol}, the primitive library is defined in the spatially separated state lattice space to enable the correspondence between sliding windows and primitive anchors. We uniformly sample $M_\theta $ polar angles and $M_\varphi $ azimuth angles (denoted by $\theta$ and $\varphi$, respectively) in the visible range of depth sensor. The motion primitive ${\bm{p}}_{ij}$ in the camera coordinate frame is expressed by endpoint position, with higher-order derivatives set to zero:
\begin{equation}
        {{\bm{p}}_{ij}} = (r\cos {\theta _j}\cos {\varphi _i},\ r\cos {\theta _j}\sin {\varphi _i},\ r\sin {\theta _j}) \,.
        \label{primitive}
\end{equation} 
Where $i\!\in\! [1, M_ \varphi]$, $j \!\in\! [1, M_ \theta]$ represent the index of primitive in horizon and vertical directions respectively, and $r$ is the radius of planning horizon.  

Similar with the state-of-the-art detectors, we predict the offsets $\Delta \theta$, $\Delta \varphi$, and $\Delta r$, as well as the end derivatives, to replace the back-end optimization in classical planners. The refined end-state of the $ij$-th trajectory can be expressed by
\begin{equation}
        \! \! \! \! {{\bm{f}}_{ij}(T)}\! = \!(r'_{ij} \cos {\theta' _{ij}}\cos {\varphi' _{ij}},\, r'_{ij}\cos {\theta' _{ij}}\sin {\varphi' _{ij}},\, r'_{ij}\sin {\theta' _{ij}}) \! \! \!
        \label{eq4}
\end{equation}
Where $r'_{ij} = r + \Delta r_{ij}$, $\varphi'_{ij} = \varphi _i + \Delta \varphi _{ij}$, and $\theta' _{ij} = \theta _j + \Delta \theta _{ij}$. By adjusting the length $r'_{ij}$, our policy can balance the speed and safety during tracking or navigation. Finally, the trajectory can be obtained by substituting the initial and predicted final derivatives into (\ref{trajectory}) for a closed-form solution.

\subsection{Unified Network} \label{Sec_Network}

\subsubsection{Network Architecture}
In this section, we unify the separate components of target tracking including perception, detection, and navigation into a single neural network. 
As shown in Fig. \ref{system_overview}, it takes the RGB-D images and 6-dimensional states (initial velocity and acceleration) as inputs, and predicts 14-dimensional outputs for each primitive, including the offsets, end-derivatives, trajectory costs, objectness scores, and target positions. The framework can easily degenerate into a more general obstacle avoidance policy by adding the goal direction as input and removing the predictions related to the target.

Specifically, we take the RGB-D images from the RealSense camera with a 90-degree horizontal FOV and a 16:9 aspect ratio as inputs, and resize them to a resolution of 160$\times$96 pixels. We divide the images into $M_\varphi  \times M_\theta $ grids, each of them corresponds to a primitive in its frustum. We modify ResNet-18 to serve as the backbone with a 32$\times$ downsampling rate for feature extraction, resulting in a $M_\varphi  \times M_\theta = 5\times 3$ feature map with 512-dimensional features. Focusing on tracking larger targets such as humans and emphasizing the design of a unified framework, we employ a simple division and general-purpose ResNet without complex tricks.

Besides, since both observations and predictions are carried out in the camera coordinate frame, only the velocities and accelerations are needed as state inputs. It is worth noting that by scaling time as $f(\alpha t)$ along each axis in (\ref{trajectory}), the flight speed can be adjusted proportionally while maintaining the geometric shape of the trajectory. This maintains the safety and smoothness of the trajectories as the speed changes. Leveraging this property, we perform inference in the normalized space, allowing the network to handle varying speeds without retraining:
\begin{equation}
        \begin{aligned}
        {{\dot f}_{n}}(t) &= {\dot f}(\alpha t)\,/\,(\alpha \, v_{max}) \\
        {{\ddot f}_{n}}(t) &= {\ddot f}(\alpha t)\,/\,({\alpha ^2} \, a_{max})\,.
        \end{aligned}
\end{equation}
Where $\alpha$ is the scaling ratio of the desired maximum speed between practice and training, $v_{max}$ and $a_{max}$ stand for the desired maximum velocity and acceleration in training, and the normalized initial derivatives $[{\dot f}_n(0), {\ddot f}_n(0)]$ is the input of network. In this way, we enhance the flexibility of the trained policy and lighten the burden of considering different desired speeds in training. Subsequently, the normalized derivatives are transformed into the local coordinate frames of $M_\varphi  \times M_\theta $ grids (defined from camera toward each grid center) to ensure the equivalence of local information for each prediction.

\begin{figure}[t]\centering
        \includegraphics[width=0.825\linewidth]{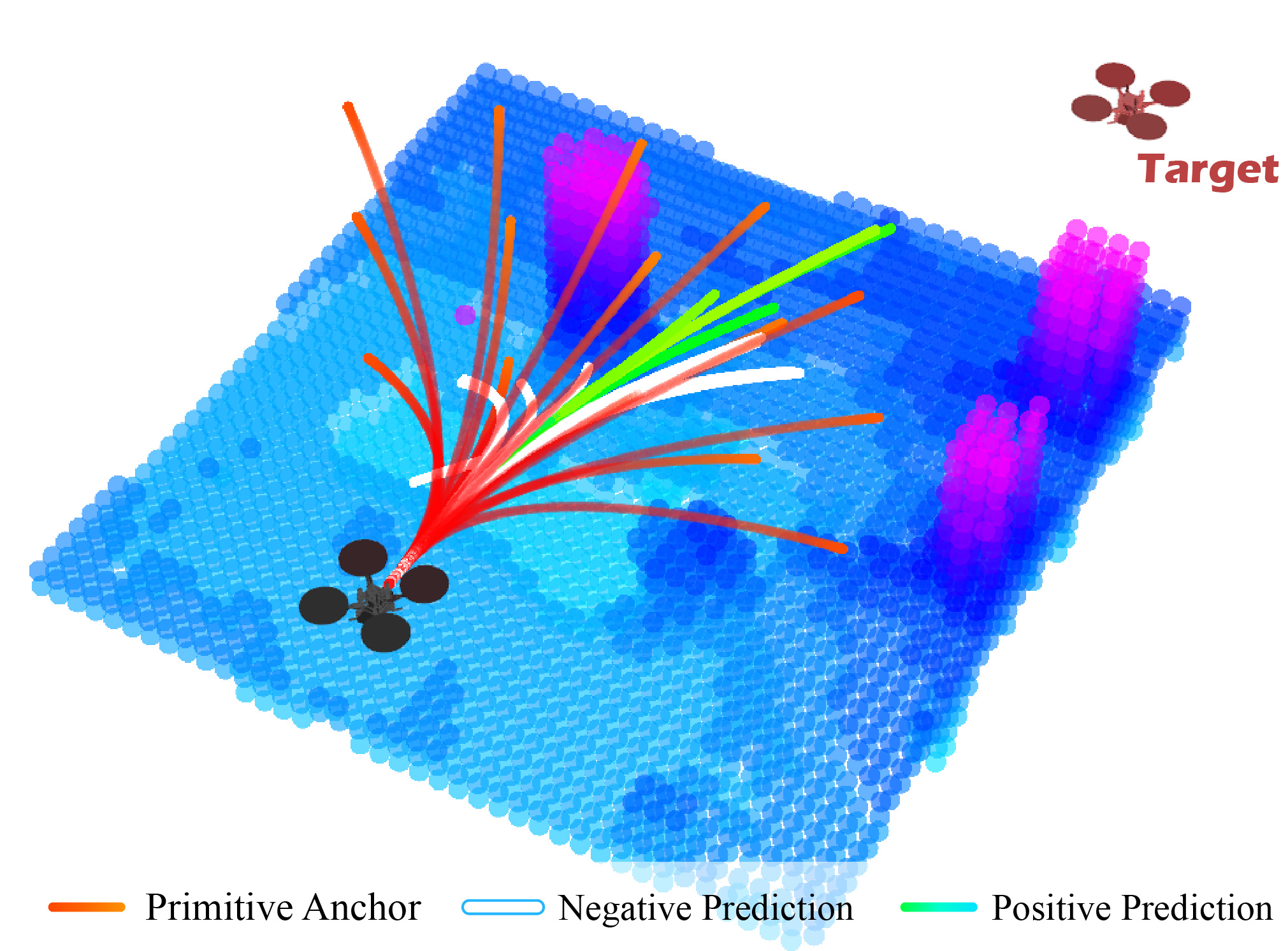}
        \caption{We predict offsets and derivatives to improve the primitives (red) and then filter out negative predictions (white) based on the objectness score. Finally, we select the best prediction from the remainings (green) based on the predicted safety and smoothness cost.}\label{pos_neg_samples}
\end{figure}

Finally,  the state inputs concatenated with image features are processed by a series of $1\times 1$ convolutional layers with shared weights to output ${\bm{y}}$ with dimensions of $M_\varphi  \times M_\theta \times 14$, including the offsets $[y_\theta, y_\varphi, y_r]$, end-derivatives in three axes $[{\bm{y}}_v, {\bm{y}}_a]$, trajectory cost $y_c$, objectness score $y_o$, and target positions $[y_{\Delta u}, y_{\Delta v}, y_d]$ for each primitive. The offsets are constrained within a frustum by a hyperbolic tangent activation function:
\begin{equation}
        \begin{aligned}
                \Delta \theta  &= \tanh({y_\theta })\,{d_\theta }\\
                \Delta \varphi  &= \tanh({y_\varphi })\,{d_\varphi} \\
                \Delta r  &= \tanh({y_r })\,{r } \,.
        \end{aligned}
        \label{eq6}
\end{equation}
Where the offset range $d_\theta$ and $d_\varphi$ are set slightly larger than the local FOV of grid, and 
we omit the primitive index subscript $ij$ for brevity. Besides, the derivative outputs ${\bm{y}}_v$ and ${\bm{y}}_a$ are also post-processed and denormalized to actual final derivatives:
\begin{equation}\
        \begin{aligned}
        {\dot{\bm{f}}(T)}  &= \tanh({ {\bm{y}}_v })\,\alpha \,{ v_{max} } \\
        {\ddot{\bm{f}}(T)}  &= \tanh({ {\bm{y}}_a })\,\alpha ^2 \, { a_{max} } \,.
        \end{aligned}
        \label{eq7}
\end{equation}
Empirically, the execution time is defined as a constant $T=2r/\alpha v_{max}$. Subsequently, the above states transformed into the camera coordinate frame are substituted into equations (\ref{trajectory}) and (\ref{eq4}) to obtain the final trajectory. In addition, the cost item $y_c$ reflects the performance of the predicted trajectory, which is explained in Section \ref{Sec_Training}.

Similar to the general detection methodologies, each grid predicts a target represented by the pixel offsets $[y_{\Delta u}, y_{\Delta v}]$ relative to the location of grid, along with the target's distance $y_d$. The center coordinate $(u, v)$ of target in pixel space can be expressed by:
\begin{equation}\
        \begin{aligned}
        u  &= (\sigma ({ y_{\Delta u} }) + u_{grid}) \,  ds \\
        v  &= (\sigma ({ y_{\Delta v} }) + v_{grid}) \,  ds \,.
        \end{aligned}
\end{equation}
Where $\sigma(\cdot)$ is the sigmoid activation function that bounds the predictions to fall between 0 and 1, $(u_{grid}, v_{grid})$ is the coordinate of grid, and $ds$ is the downsampling rate of the network which is set to 32$\times$ here. Given the current camera pose ${}^w\bm{T}_{c} \in SE(3)$ by state estimator module, the target location in world coordinate frame can be projected by 
\begin{equation}\
        \bm{p}_w = {}^w\bm{T}_{c} \, \bm{C}^{-1} y_d\, \bm{p}_{uv}\,.
        \label{eq_9}
\end{equation}
Where $\bm{p}_{uv}=[u,v,1]^T$ and $\bm{p}_w=[x,y,z]^T$ are the target position in pixel and world space respectively, and $\bm{C} = [f_x, 0, c_x;\, 0, f_y, c_y;\, 0, 0, 1]$ is the intrinsic matrix of camera. Finally, the objectness score is also processed through a logistic activation to handle the binary classification task of target \textit{vs}. background.

\subsubsection{Tracking Strategy}

\begin{figure}[t]\centering
        \includegraphics[width=\linewidth]{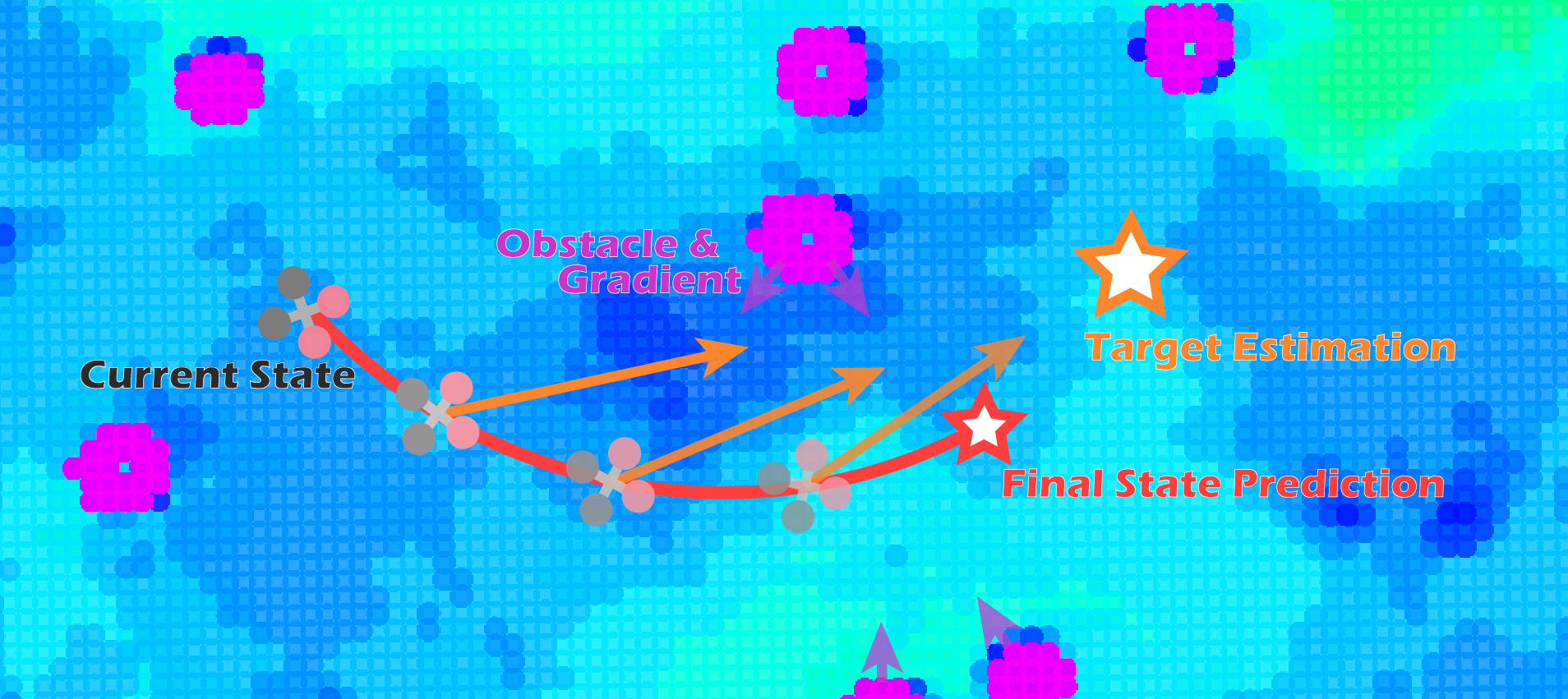}
        \caption{We predict the target's position separately, since the final state usually doesn't coincide with the target due to the combined cost. Subsequently, the trajectory is discretized for execution, and the yaw is always planned towards the target.}\label{yaw_planning}
\end{figure}

We simultaneously predict the states of $M_\varphi  \times M_\theta$ trajectories in one propagation, but during application, only the optimal one is solved for execution. To reduce redundancy, we adopt non-maximum suppression (NMS) on the predictions based on their costs, objectness scores, and spatiotemporal consistency of the detection. As visualized in Fig. \ref{pos_neg_samples}, we first filter out the invalid predictions that do not contain the target based on the objectness scores. Subsequently, we predict and update the target position using extended Kalman filtering (EKF) under the assumption of constant velocity. This does not conflict with the unpredictability of the target, as it is only used for state estimation with a large process noise matrix to filter out false positives and avoid missed detection, rather than predicting the future trajectory of the target. We perform the regular prediction step at a fixed frequency, but in the update step, we first attempt to update the state $\bm{x}$ with all remaining detections to temporary variable $\bm{x}_{temp}$:
\begin{equation}
        \bm{x}_{temp} = \bm{x} + \bm{K}({\bm{p}_w} - \bm{Hx})\,.
\end{equation}
Where $\bm{K}\in \mathbb{R}^{3\times 6}$ is the latest Kalman gain, $\bm{H} = [\bm{I, 0}]\in \mathbb{R}^{3\times 6}$ is the measurement matrix, and $\bm{x}$, $\bm{x}_{temp}$ are the state vector including position and velocity. Based on the spatiotemporal continuity of the target, we filter out the false positives with the inconsistency $|\bm{x}_{temp} - \bm{x}|$ greater than a threshold, and perform actual update of $\bm{x}$ using the detection result with the smallest inconsistency below the threshold. For predictions that remain valid, we select the optimal one based on the smoothness and safety cost $y_c$, and then substitute corresponding offsets into equations (\ref{trajectory}) and (\ref{eq4}) to solve the trajectory.

Since yaw is independent in the differential flatness space, we plan it separately. As illustrated in Fig. \ref{yaw_planning}, we discretize the trajectory for execution and ensure that the desired yaw is always oriented towards the estimated target, which is fused using EKF to ensure smoothness. For frames without target detected, we only perform prediction step of EKF and orient the yaw towards the direction in which the target is lost. Through the extremely straightforward end-to-end design, we achieve instantaneous tracking responses within milliseconds, which is one order of magnitude faster than traditional pipelines. As evidenced by a series of recent breakthroughs in high-speed flights \cite{science, Newton} and experiments in Section \ref{Sec_Experiment}, the processing latency is often more important, especially on compact quadrotors with limited visual sensors and onboard computational resources.

\subsection{Control Strategy} \label{Sec_Contorl}

\begin{figure}[t]\centering
        \includegraphics[width=\linewidth]{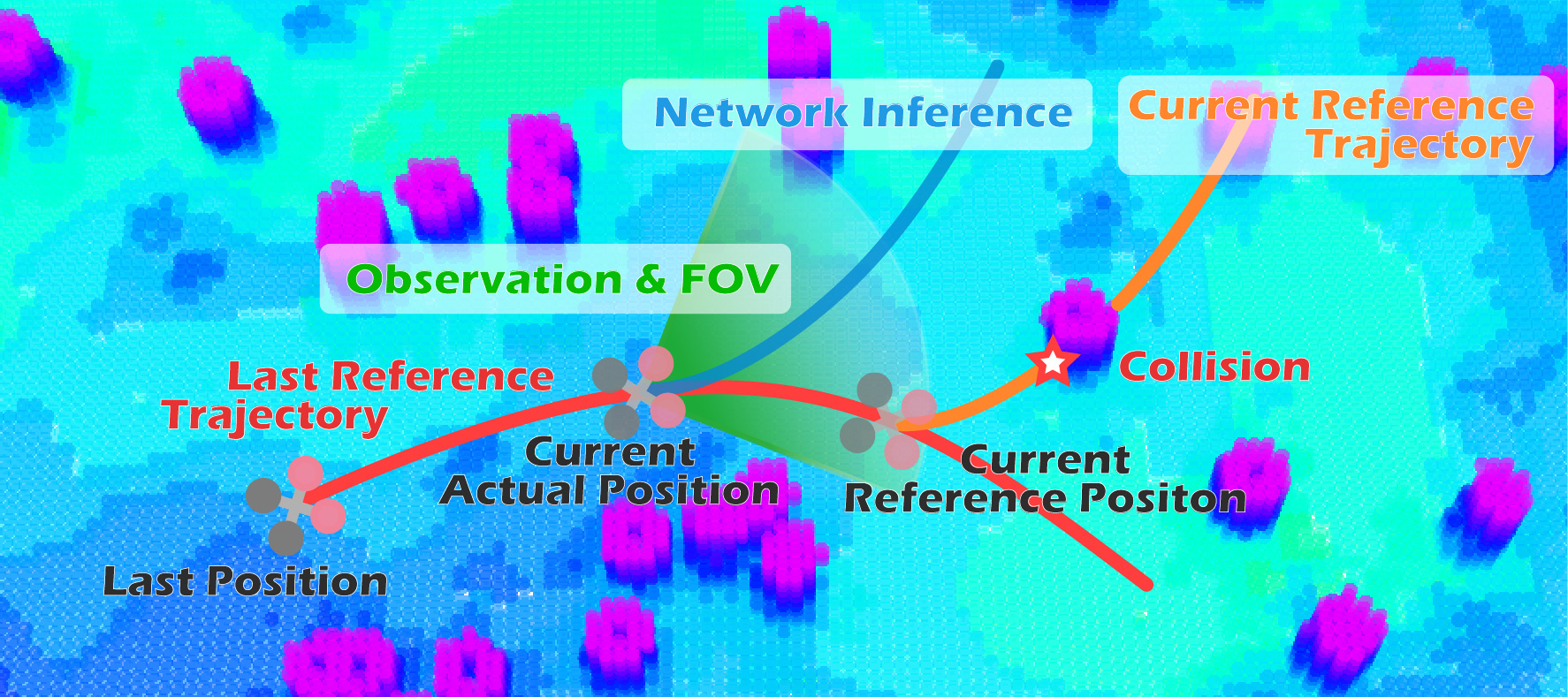}
        \caption{Planning in differential flatness space and tracking by position controller need to plan from the reference position. However, since the image is observed at actual position (green sector), it might be fatal to the mapping-free end-to-end planners with the control errors increasing.}\label{plan_from_reference}
\end{figure}

The differential flatness property of the quadrotors enables independent trajectory planning in $\{x,y,z\}$ directions using polynomials or B-splines, which are then tracked by the position controller. To ensure the continuity and smoothness of the reference states in control, the planners take the previous reference state as initial value for receding-horizon optimization, rather than the current actual state. This is reasonable in mapping-based approaches \cite{fast_planner} \cite{ego_planner}, as the environmental information at the reference position is contained in map. However, for the end-to-end policies such as \cite{science, yopo}, it can be fatal with the tracking error between reference and actual states increasing in high-speed flight, because the images are only observable at the actual position. As shown in Fig. \ref{plan_from_reference}, the trajectory inferred by the network with the current observation (the green sector) is safe under the actual situation (the blue trajectory), but would lead to collision if planned from the reference position (the orange one). Moreover, the cascade control structure, which connects the outer loop (position control) and the inner loop (attitude control) in series, will lead to error accumulation and response latency, making aggressive flight challenging. 

To avoid this inconsistency and decrease the error accumulation in hierarchical control, we further remove the position controller and directly plan the attitude from higher-order derivatives of the trajectory, thereby eliminating the continuity constraint on reference position. Assume the quadrotor model is a 6 degree-of-freedom rigid body with four individual rotors, the dynamic of quadrotor can be described by:
\begin{equation}
        \ddot {\bm{f}} =  \frac{1}{m}{}^w\!\bm{R}_bF_c{\bm{e}_z} - \bm{g} + \frac{1}{m}\bm{d}\,.
        \label{dynamic}
\end{equation}
Where $\ddot {\bm{f}}$ is the acceleration of quadrotor in the world frame, $m$ is the mass, ${}^w\!\bm{R}_b \in SO(3)$ is the attitude, $F_c$ represents the collective thrust, ${\bm{e}_z}=[0, 0, 1]^T$ means the z-direction of the body frame, $\bm{g} = [0, 0, 9.8]^T$ donates the gravity, and $\bm{d}$ is the lumped disturbance which is primarily caused by aerodynamic drag force and external disturbances. Therefore, by specifying the desired acceleration $\ddot{\bm{f}}(t)$ through the network planner and assuming that the disturbances $\bm{d}(t)$ can be estimated online, we can compute the desired attitude ${}^w\!\bm{R}_b = [\bm{R}_x, \bm{R}_y, \bm{R}_z]$ of the quadrotor by leveraging differential flatness property:
\begin{equation}
        \begin{aligned}
                {\bm{R}_z} &= \frac{\bm{F}}{{\left\| \bm{F} \right\|}}\\
                {\bm{R}_y} &= \frac{{{\bm{R}_z} \times {\bm{R}_{yaw}}}}{{\left\| {{\bm{R}_z} \times {\bm{R}_{yaw}}} \right\|}}\\
                {\bm{R}_x} &= {\bm{R}_y} \times {\bm{R}_z}\,,
        \end{aligned}
\end{equation}
where the column vector $\bm{F}$ and ${\bm{R}_{yaw}}$ can be calculated by:
\begin{equation}
        \begin{aligned}
        &\bm{F} = [\ddot{f_x} - {d_x},\,\ddot{f_y} - {d_y},\,\ddot{f_z} - {d_z} + g]^T\\
        &\bm{R}_{yaw} = {[\cos (yaw),\,\sin (yaw),\,0]^T}\,,
        \end{aligned}
        \label{dynamic_end}
\end{equation}
and the desired collective thrust can be obtained by $F_c=m||\bm{F}||$.

To compensate for the mismatch between the actual flight and network-processed point-mass model, we introduce a high-gain disturbance observer to estimate the lumped disturbances. Consider the dynamic (\ref{dynamic}) which can be expressed in the following multivariable first-order system:
\begin{equation}
        \dot{\bm{z}_1} = {}^w\!\bm{R}_b F_c \bm{e}_z - m\bm{g} + \bm{d}\,
\end{equation}
where $ {\bm{z}_1} = m\bm{v}$. Then, the high-gain disturbance observer can be designed as:
\begin{equation}
        \begin{aligned}
                &{\dot{\hat {\bm{z}}}_1} = {}^w\!\bm{R}_b F_c \bm{e}_z - m\bm{g} + {{\hat {\bm{z}}}_2} + {\frac{{{\alpha _1}}}{\zeta }}({\bm{z}_1} - {{\hat {\bm{z}}}_1})\\
                &{\dot{\hat {\bm{z}}}_2} = {\frac{{{\alpha _2}}}{\zeta ^2 }}({\bm{z}_1} - {{\hat {\bm{z}}}_1})\,.
        \end{aligned}
\end{equation}
Where positive constants $\alpha _1$, $\alpha _2$, and  $\zeta$ (where $\zeta \ll 1$) determine the sensitivity and stability of the disturbance observer, ${\hat {\bm{z}}}_1$ and ${\hat {\bm{z}}}_2$ are the estimation of $\bm{z}_1$ and $\bm{d}$ respectively. Relying solely on the current speed $\bm{v}$ provided by the visual odometry and the last control commands ${}^w\!\bm{R}_b F_c$, we can recursively estimate the current disturbance. Subsequently, the disturbance and the second derivative of the predicted trajectory are substituted into equations (\ref{dynamic})-(\ref{dynamic_end}) to solve the desired thrust and attitude, which are then used as inputs of the autopilot. 

Compared to the state-of-the-art control-based policies, our network is decoupled from external disturbances and analytical dynamics, eliminating the need for precise model identification or domain randomization through disturbance injection during training. We simplify the real-world deployment as the policy is independent of the actual physical model and inputs of the disturbance observer can be directly accessed from the state estimator. Furthermore, compared to instantaneous control commands, predicting trajectories in differentially flat space naturally ensures smoother flight and facilitates the tight integration of deep learning with traditional optimization-based planners, enabling end-to-end gradient descent using the trajectory costs.

\subsection{Training Strategy}  \label{Sec_Training}
In this work, the formulation of the cost function for each prediction is expressed by:
\begin{equation}
        \begin{aligned}
        \! \! \!\mathcal{L} &= {[\mathds{1} ^{pos}({\lambda _s}{J_s} + {\lambda _c}{J_c} + {\lambda _g}{J_g} + {\mathcal{L} _{cost}} + {\mathcal{L} _{tgt}} + {\mathcal{L} _{obj}})}  \\
        &+ {\lambda _1}(1 - \mathds{1}^{pos})({\lambda _s}{J_s} + {\lambda _c}{J_c} + {\mathcal{L}_{cost}}) + {\lambda _2}\mathds{1}^{neg}{\mathcal{L}_{obj}}]
        \end{aligned}
        \label{total_cost}
\end{equation}
Where $\{J_s, J_c, J_g\}$ are smoothness, safety, and goal cost of the predicted trajectory, $\{\lambda_s, \lambda_c, \lambda_g\}$ are the weighing terms for trajectory optimization, and $\mathcal{L} _{cost}$ is the loss of predicted trajectory cost. Besides, $\{\mathcal{L} _{tgt}, \mathcal{L} _{obj}\}$ are the loss of predicted target position and objectness score, respectively. The binary variable $\{\mathds{1}^{pos}, \mathds{1}^{neg}\}$ indicates whether this primitive is positive or negative sample, while $\{\lambda_1=0.2,\lambda_2=0.5\}$ are used to balance the positive, negative, and ignored samples. By ignoring the loss $\{\mathcal{L} _{tgt}, \mathcal{L} _{obj}\}$ associated with detection and treating all predictions as positive samples, the cost function can degenerate to general navigation tasks. The definitions of the aforementioned costs and sample partitioning are explained in detail in the following sections.

\subsubsection{Trajectory Optimization} To train the network policy in autonomous navigation, reinforcement learning and imitation learning are widely employed by either imitating a privileged expert or exploring the optimal policy through trial-and-error. In actor-critic-based reinforcement learning, the critic network is trained to evaluate actions and guide the actor network's gradient ascent. However, in navigation tasks, it can be replaced by the differentiable environment (such as ESDF) which is able to provide more direct and accurate evaluation with numerical gradients. In imitation learning, the network is trained by decreasing the distance between predictions and labels, which can be generated by a privileged gradient-based expert. However, the gradients can be directly applied to the weights of network through the chain rule, without the need to first optimize the expert via multi-step gradient descent.
Based on this, we develop an end-to-end training method that directly back-propagates gradients of trajectory cost to the network to guide the training process. 

Firstly, to avoid jerky maneuvers caused by temporal inconsistent predictions and infeasible dynamics due to excessive curvature, we follow \cite{richter_gradient} to constrain the smoothness cost as the integral of squared derivatives along the trajectory. With the form represented by equation (\ref{trajectory}), the smoothness cost can be expressed as:
\begin{equation}
        {J_s} = \bm{dM}^T\bm{QMd}\,.
\end{equation}
Where the Hession matrix $\bm{Q}$ is constructed from the integral of the square higher-order derivatives of the polynomial. Denote $\bm{M}^T\bm{QM}$ as matrix ${\bm{B}}$, then the cost can be written in a partitioned form as:
\begin{equation}
        {J_s} = {\left[ {\begin{array}{*{20}{c}}
                {{{\bm{d}}_F}}\\
                {{{\bm{d}}_P}}
                \end{array}} \right]^T}
                        \left[ {\begin{array}{*{20}{c}}
                        {{{\bm{B}}_{FF}}}&{{{\bm{B}}_{FP}}}\\
                        {{{\bm{B}}_{PF}}}&{{{\bm{B}}_{PP}}}
                        \end{array}} \right] 
                \left[ {\begin{array}{*{20}{c}}
                {{{\bm{d}}_F}}\\
                {{{\bm{d}}_P}}
                \end{array}} \right]\,.
\end{equation}
The gradient of $J_s$ with respect to the final state ${\bm{d}}_P$ can be computed as:
\begin{equation}
        \frac{{\partial {J_s}}}{{\partial {{\bm{d}}_P}}} = 2{\bm{d}}_F^T{{\bm{B}}_{FP}} + 2{\bm{d}}_P^T{{\bm{B}}_{PP}}\,.
\end{equation}

Subsequently, to avoid collisions in obstacle-dense environments during tracking or navigation, we reformulate the safety cost in \cite{gao_gradient} as the time integral of the potential function $c(\cdot)$, since we fix the execution time $T$ and treat the endpoint as free variable:
\begin{equation}
        \begin{aligned}
                {J_c} &= \int_0^T {c(\bm{f}(t))} dt\\
                 &= \sum\limits_{\kappa   = 0}^{T/\delta t} {c(\bm{f}({\kappa \cdot \delta t }))\,\delta t} \,.
        \end{aligned}
        \label{Jo}
\end{equation}
Where the potential function is expected to increase rapidly as the quadrotor approaches the obstacle and flatten out as it moves away (e.g., exponential function with respect to the negative distance to obstacle). Then, the gradient of $J_c$ in discrete form is:
\begin{equation}
        \frac{{\partial J_c}}{{\partial {{\bm{d}}_{P}}}} = \sum\limits_{\kappa  = 0}^{T/\delta t} {{\nabla }c(\bm{f}(\kappa \cdot \delta t ))} {\bm{T}_\kappa}{{\bm{L}}_P} \,\delta t \,.
\end{equation}
Where ${\bm{T}_\kappa} = [(\kappa \cdot \delta t) ^ 0, (\kappa \cdot \delta t) ^1, ..., (\kappa \cdot \delta t) ^{5}]^T$, ${\nabla }c(\cdot)$ is the gradient of the potential function along axis which can be queried from the ESDF, and ${\bm{L}}_P$ is the right block of matrix ${{\bm{M}}}$ which corresponds to the final state. 

Finally, to track the moving target (in tracking tasks) or guide the quadrotor flying towards the manually specified goal (in navigation tasks), we define the goal cost as follows:
\begin{equation}
        {J_g} = {({\bm{d}_{Pp}} - \bm{g}_p)^2} \,.
\end{equation}
Where $\bm{d}_{Pp}$ is the position component of ${{\bm{d}}_{P}}$, and $\bm{g}_p$ is the position of the target or goal. To ensure numerical stability, we normalize the goal to a direction vector with fixed length in navigation tasks, as the goal may vary from meters to hundreds of meters away. In contrast, in tracking tasks, the target's position is not scaled, which ensures longer trajectories for faster pursuit when target is distant, while shorter ones for deceleration as the target is approached. Then, the gradient of ${J_g}$ can be simply calculated as:
\begin{equation}
        \frac{{\partial {J_g}}}{{\partial {{\bm{d}}_{Pp}}}} = 2({{\bm{d}}_{Pp}} - \bm{g}_p) \,.
\end{equation}

After that, the gradients of network's predictions can be obtained via the chain rule:
\begin{equation}
        \begin{aligned}
                \frac{{\partial J_\xi }}{{\partial {y_\epsilon  }}} &= \frac{{\partial J_\xi}}{{\partial {{\bm{d}}_P}}} \, \frac{{\partial {{\bm{d}}_P}}}{{\partial \Delta \epsilon }} \, \frac{{\partial \Delta \epsilon }}{{\partial {y_\epsilon  }}} \\
                \frac{{\partial J_\xi}}{{\partial {\bm{y}_\varepsilon  }}} &= \frac{{\partial J_\xi}}{{\partial {{\bm{d}}_P}}} \, \frac{{\partial {{\bm{d}}_P}}}{{\partial {\bm{y}_\varepsilon }}} \,.
        \end{aligned}
        \label{chain_rule}
\end{equation}
Where $\xi \in \{s, c, g\}$, $\epsilon \in \{\theta,\varphi , r\} $, and $\varepsilon \in \{v, a\}$. The final results can be calculated by substituting (\ref{eq4}), (\ref{eq6}), and (\ref{eq7}). Note that above costs and gradients are computed independently along each axis $\mu \in \{x, y, z\}$ and we omit the subscript for simplicity. 

Besides, the predicted cost $y_c$ is supervised by the trajectory cost $\hat y_c$ with the smooth L1 loss:
\begin{eqnarray}
        \mathcal{L}_{cost} = Smooth L_1(\hat y_c, y_c)\,.
\end{eqnarray}
Where $\hat y_c={\lambda _s}{J_s} + {\lambda _c}{J_c} + {\lambda _g}{J_g}$ in navigation tasks, whereas $\hat y_c={\lambda _s}{J_s} + {\lambda _c}{J_c}$ in tracking tasks since target may not exist. It is analytic and can be automatically derived by the deep learning frameworks. 

\begin{figure}[t]\centering
        \includegraphics[width=0.9\linewidth]{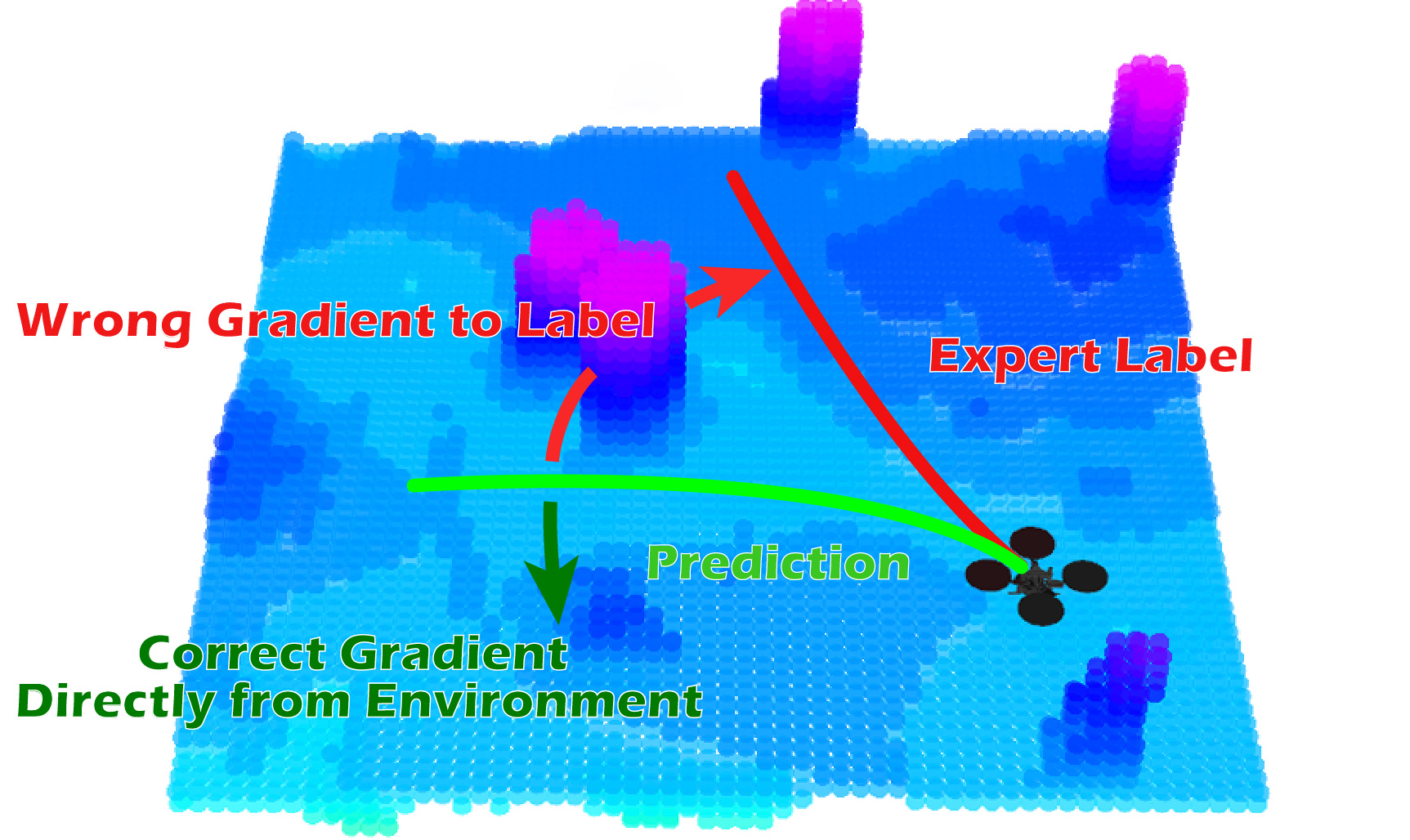}
        \caption{Due to the multimodal nature of navigation, the distance to limited expert demonstrations, even with WTA loss, cannot accurately reflect the performance. In contrast, we directly back-propagate the actual gradients from environment to the network, which is more accurate and straightforward.} \label{gradient_comparison}
\end{figure}

Compared to stochastic exploration of the policy in reinforcement learning, the direct gradient descent from trajectory cost and differentiable environment improves data utilization and convergence efficiency, without the need for constant interaction with the simulator and time-consuming online sensor rendering. As visualized in Fig. \ref{gradient_comparison}, in comparison with the distance to expert demonstration in imitation learning, the proposed cost function can accurately represent the prediction's performance, especially in multimodal problems. Furthermore, the exclusion of the expert policy and label assignment strategy makes the training process fully end-to-end. 

\subsubsection{Target Detection}

In tracking tasks, due to the combined constraints of smoothness and safety cost, the endpoint of the trajectory may not coincide with the target. Therefore, in addition to optimizing the trajectory to closely track the target, we also independently predict the target's position for yaw planning to ensure the quadrotor heading towards the target. Similarly with equation (\ref{eq_9}), the target's position in camera coordinate frame can be expressed by $ \bm{p}_c = \bm{C}^{-1} y_d\, \bm{p}_{uv}$. Assuming the ground truth of target's relative position $\hat {\bm{y}}_c$ is accessible during training, the loss function of detection can be expressed by:
\begin{eqnarray}
        \mathcal{L}_{tgt} = Smooth L_1(\hat {\bm{p}}_c, \bm{p}_c)\,.
\end{eqnarray}

In this work, we treat the detection task as a simplified binary classification problem, and assign a binary objectness label (target or background) to each anchor. We follow the object detection framework, categorizing all predictions into positive samples, negative samples, and ignored samples. In detail, we project the target's position $\hat {\bm{y}}_c$ onto the image plane according to perspective projection. The grid containing the projected target is defined as the positive sample and the corresponding label $\hat y_o$ is set to 1. Grids far from the target are treated as negative samples and assigned label of 0. Otherwise, grids close to the target are marked as ignored samples to avoid ambiguity during training. Finally, the objectness score is trained by binary cross-entropy (BCE) loss:
\begin{eqnarray}
        \mathcal{L}_{obj} = BCE(\hat {y}_o, \sigma (y_o))
\end{eqnarray}

In conclusion, as expressed in equation (\ref{total_cost}), all cost functions are applied to positive samples for tracking and detection. For negative and ignored samples, we optimize the safety and smoothness costs to ensure the feasibility of trajectory in testing when no target appears in this gird or missed detection. The trajectory costs of all primitives are trained for prediction selection, while the objectness scores for ignored samples are not considered to avoid confusion. 

\subsubsection{Training Dataset}

Leveraging privileged information and the data-driven capability of deep learning, the network policy can achieve robust navigation and tracking performance using only noisy sensor observations. Specifically, the privileged information, including the ground truth of environment (point cloud and ESDF map) and target states, is accessible for costs and gradients computation in training, while only the limited RGB-D images are available for the network. Compared to the sequential dependency in reinforcement learning, the proposed training method does not require online interaction with the simulator or expert annotations, greatly simplifying the dataset collection. This allows us to collect training data (only poses, images, and point cloud map) in advance by extensively randomizing the quadrotor's poses, thereby saving the time of online simulator rendering during training and enhancing scalability to real-world collection. Additionally, we randomly place target within the quadrotor's FOV and transform it into the camera coordinate system for annotation. We utilize raycasting to verify the visibility of target and prevent incorrect labeling. In real-world tracking applications, we use high-precision LiDAR to construct the ground truth of environment and record the quadrotor's state, which can be easily achieved by running LiDAR odometry. In the meantime, an RGB-D camera is used to collect images. The targets are annotated by the state-of-the-art detector YOLO and then transformed into the camera coordinate frame by combining the range information. 

\begin{figure}[t]\centering
        \includegraphics[width=\linewidth]{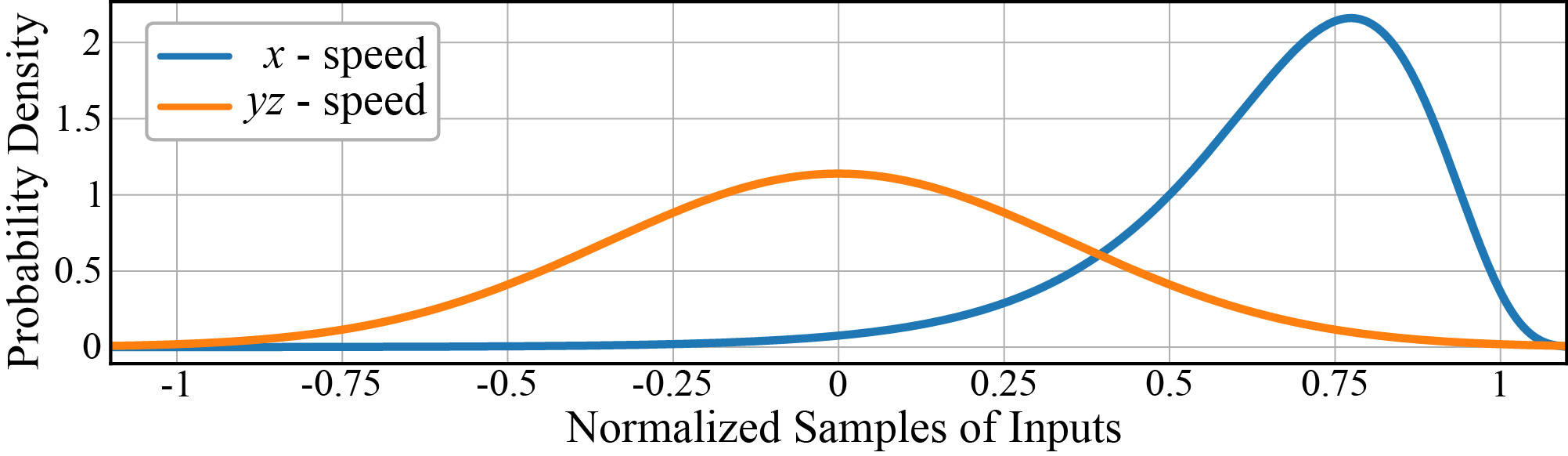}
        \caption{Probability density function of state sampling.}\label{speed_sampling}
\end{figure}

For better generalization, we perform data augmentation on RGB observations in the HSV (hue, saturation, and value) color space. Additionally, we randomly sample state observations (velocity and acceleration) and assign varying states to every image in each training step to improve the utilization efficiency of image data. As shown in Fig. \ref{speed_sampling}, $yz$-velocity and acceleration are sampled from a normal distribution based on actual measurements, while the forward velocity follows a log-normal distribution as:
\begin{eqnarray}
        f(v) = \frac{1}{{({v_m} - v)\sqrt {2\pi \sigma } }}\exp \left[ { - \frac{1}{{2{\sigma ^2}}}{{(\ln ({v_m} - v) - \mu )}^2}} \right]
\end{eqnarray}
where $\sigma$ and $\mu$ are the parameters of distribution, $v_m$ is slightly larger than the maximum velocity to ensure that the samples can encompass the maximum speed. This ensures that the data remains diverse throughout multiple epochs of training. For imitation learning, such data augmentation means that each random sample requires dozens of gradient descent steps to optimize the expert label, whereas each label updates the network only one step during training, leading to redundant computation. Therefore, the proposed training method not only accurately reflects the performance of the trajectory but also improves data efficiency.

\section{Experiment}  \label{Sec_Experiment}
In this section, we demonstrate the performance of the proposed tracking and navigation framework in both simulation and real-world scenarios. By contrast with traditional pipelines, we illustrate the efficiency of the proposed end-to-end tracker, as well as its capability for high-speed tracking in cluttered environments. Furthermore, to validate the scalability of proposed algorithm, we conduct a series of experiments on high-speed navigation in obstacle-dense scenarios. More experimental details can be found in the supplementary video.

\subsection{Quadrotor Platform}

\begin{figure*}[t]\centering
        \includegraphics[width=\linewidth]{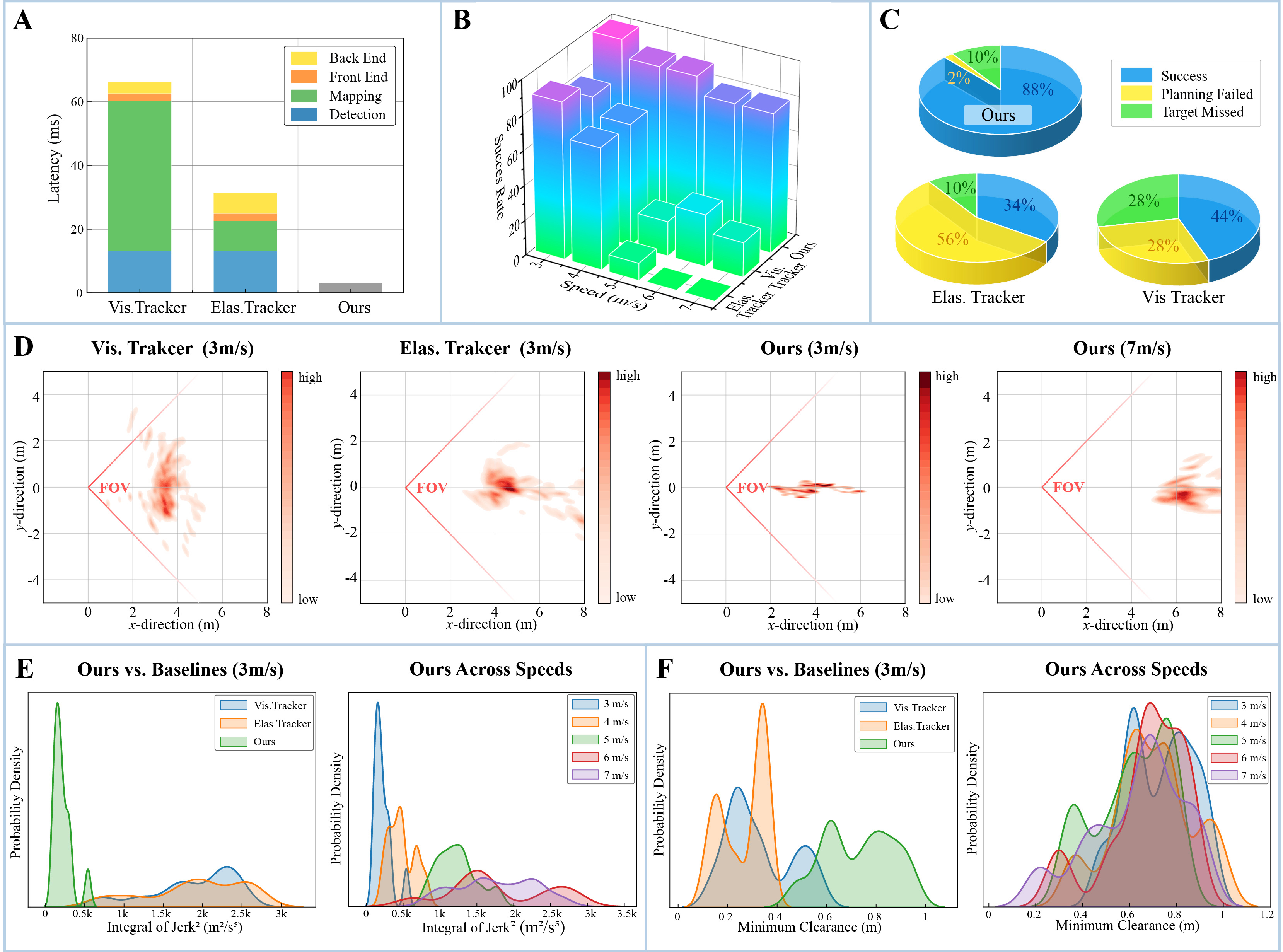}
        \caption{Tracking comparison of real-time performance, success rate, visibility, smoothness, and safety metrics. (A) Real-time performance comparison with the baselines, where the latency of our end-to-end method is marked in gray. (B) Success rate comparison with the baselines at different tracking speeds. (C) Performance analysis summarized across all tests, where planning failed refers to collision or emergency stop, while target missed occurs when the target escapes the FOV or is occluded. (D) The distribution of  target positions relative to the tracker on x-y plane, where the sector angle represents the FOV of quadrotor. (E) Smoothness cost comparison with the baselines and evaluation of  proposed method at different speeds. (F) Safety clearance comparison with the baselines and evaluation of our method at different speeds.}\label{tracking_compare}
\end{figure*}

In this section, we design a high-speed aerial robot to fulfill the tasks of agile tracking and navigation in unknown environments relying solely on the onboard sensing and computation. As visualized in Fig. \ref{cover}, the airframe is designed based on OddityRC XI35 FPV frame and equipped with OddityRC-2006 2150KV motors, each providing over 1000 g of thrust with GEMFAN 3.5-inch propellers. The drone is designed with a compact structure, featuring a wheelbase of 155 mm and total weight of 850 g. It achieves the max thrust-to-weight ratio of 4.7 in theory, which enables agile flights in cluttered environments. We utilize a commonly adopted FPV flight tower, consisting of a NxtPX4 flight controller and a HobbyWing XRotor 4-in-1 electronic speed controller. We employ the open-source firmware PX4 for attitude control, which offers better flexibility and scalability. Besides, the RealSense D455 camera is utilized to provide RGB-D images with the 16:9 aspect ratio and horizon FOV of 90°. Compared to high-precision LiDAR, visual sensor is smaller, lighter, and cheaper, enabling more compact aircraft design. More importantly, it can provide high-frame-rate image observations with rich textures and colors, enabling agile tracking in cluttered environments. However, compared to LiDAR with effective range exceeding hundreds of meters, depth camera has an extremely limited perception range of less than 10 meters and suffers from considerable noise, which poses great challenge to the efficiency and robustness of the planner. Besides, a lightweight NVIDIA Orin NX (weighing 180 g) is used as the onboard computer, which has an 8-core Cortex-A78 CPU running at 2.0 GHz and a GPU with computing power of 100 TOPS. Thanks to the simple fully convolutional architecture and the widespread application of deep learning, the proposed algorithm can be easily deployed to edge computing platforms for acceleration (e.g., by TensorRT on NVIDIA devices), achieving immediate response in milliseconds. Besides, the state estimation is performed by the visual-inertial odometry VINS-Fusion and the reference thrust attitude are transmitted to the flight controller via MAVROS. The platform only receives a start and stop command from the ground computer, and is therefore completely autonomous during flight. The hardware is open-source and available on \href{https://github.com/TJU-Aerial-Robotics/YOPO-Tracker}{https://github.com/TJU-Aerial-Robotics/YOPO-Tracker}.

\subsection{Tracking Tasks}

\subsubsection{Simulation Comparison}

In this section, we compare the proposed approach with state-of-the-art open-source trackers \cite{elastic-tracker} (abbreviated as Elas. Tracker) and  \cite{visibility-tracker} (abbreviated as Vis. Tracker), which achieve remarkable performance at relatively lower speeds. They adopt completely different optimization frameworks with hard constraints (restricting the trajectory to the corridor and observation sector by solving constrained optimization problem) and soft constraints (optimizing trajectory by combining the gradients of safety and visibility cost), respectively. Most previous works, such as  \cite{elastic-tracker, visibility-tracker}, do not consider 3D detection and state estimation of the target but instead broadcast the target's position to tracker. However, the target's position is usually unavailable in real-world applications, as the target may be uncooperative or even adopt evasive maneuvers to avoid being tracked. Therefore, the experimental conditions are set to be more challenging in this work, where only limited onboard sensor observations (RGB-D images) are available to simulate unforeseen occlusion and loss of target. For fairness, the widely-used detector YOLOv5-s trained on the same dataset as our method, incorporating depth information for 3D position estimation, is employed to provide target information for the baselines. Moreover, unlike structured artificial environments, our experiments are conducted in a random cluttered forest \cite{science} with an average clearance of 4 m. We use Fast Planner \cite{fast_planner} for evader which has access to the privileged point cloud map of environment to prevent planning failures during high-speed flight. In comparison, the baseline methods map online since only noisy depth images are available, making it challenging for high-speed flight as discussed in \cite{science}. Besides, the odometry of tracker is directly obtained from the simulator, as localization is not considered in this work.

We first compare the real-time performance with the baseline methods, as shown in Fig. \ref{tracking_compare}A. The testing system is Ubuntu 20.04, with the hardware of Intel i7-9700 CPU, NVIDIA RTX 3060 GPU, and 32 GB of RAM. In Fig. \ref{tracking_compare}A, mapping includes raycasting, grid map construction, and ESDF generation (if present). The front end consists of path search and corridor generation (if present), while the back end represents trajectory optimization. 
With a total computation time of 66.7 ms in serial, Vis. Tracker incurs the highest processing latency, where nearly 50 ms is spent on mapping and ESDF construction. For clarification, the mapping delay depends not only on the update range (set to $16 \times 16 \times 8 $ m$^2$ with a resolution of 0.2 m in our experiments) but also on the complexity of the environment, as ESDF are computed from all occupied grids. Therefore, our results align with those in \cite{science} since we use the same unstructured forest scenario. By contrast, the ESDF-free method Elas. Tracker demonstrates faster real-time performance. However, low-precision vision sensor and dynamic objects introduce additional noise to the map, which is particularly challenging to corridor-based method \cite{elastic-tracker}, as the flight corridor must remain clear. Therefore, temporal filtering operations are necessary to cope with perception errors. As a result, multiple observations are required to completely add an obstacle to the map, increasing the effective latency of the system further. In comparison, our method is over an order of magnitude faster than the baselines. It is able to map the sensor observations to low-level control commands within only 3 ms on average, including 2 ms for network inference after deployment on TensorRT and an additional 1 ms for post-processing, specifically. Firstly, we replace the perception and mapping processes with a lightweight network, and leverage privileged learning and data-driven nature to achieve competitive noise robustness. Secondly, inspired by the multimodal similarity between detection and navigation, we integrate both tasks into a unified network and consider the tracking criterion in trajectory prediction. Thirdly, we explore the solution space comprehensively through a set of predefined primitives (similar to the prior anchor boxes in detection), and predict offsets in parallel for further improvement. Additionally, we select the optimal trajectory based on the predicted costs and only solve its coefficients for execution. Leveraging the network's parallel prediction capability, our method avoids the linear increase of computational cost with the number of optimized trajectories in traditional topology-guided methods \cite{Topological, Tgk-planner}. Furthermore, we directly convert the prediction into attitude thrust commands, reducing the latency while eliminating error accumulation in the cascaded control structure. Through specific design, our method significantly improves the real-time performance and maintains the interpretability of system in the meantime. 

We subsequently compare the success rates of the proposed method and the baselines with the target's escape speed varies between 3 and 7 m/s. We repeat 10 experiments for each speed, initializing from different positions in a forest where trees are randomly placed according to a homogeneous Poisson point process with intensity of 1/16 tree/m$^2$. Both the tracker and target take off from unobstructed areas to ensure that the target is visible at the beginning. The goal of the target, which is unknown to the tracker, is set 40 meters ahead and randomly switched midway to simulate evasive maneuvers during escapes. It is considered successful if the tracker effectively follows the target to the endpoint without any collisions. Besides, as mentioned above, only the limited onboard visual observations are accessible to the tracker, while the ground-truth map of the environment is available to the evader. Moreover, benefiting from the normalized inputs and predictions, we used the same pre-trained model across experiments with different speeds without fine-tuning. As summarized in Fig. \ref{tracking_compare}B. The traditional baselines \cite{elastic-tracker,visibility-tracker} perform excellently at lower speeds; however, they experience significant performance drops as the target speed increases. This aligns with the experiments demonstrated in the original papers, in which the maximum speeds of the target are 2 - 2.5 m/s. Furthermore, we investigate the failure reasons during tracking, which are categorized as planning failed (collision or emergency stop) and target missed (escaping the FOV or occluded). As shown in Fig. \ref{tracking_compare}C, the failure of Elas. Tracker is primarily caused by planning failed, which can be attributed to the noise-sensitive corridor and low-precision vision sensors. As demonstrated by the recent breakthroughs, the hard-constraint-based methods \cite{ren2025safety,Bubble} exhibit remarkable performance in high-speed navigation, while relying on large-scale and high-precision LiDAR. However, the map is typically delayed and noisy with limited vision sensors and unstructured scenarios, leading to fragmented and narrow flight corridors and target observation sectors. Despite efficient back-end optimization, inappropriate constraint will lead to unreasonable solutions or planning failures. This becomes more critical at higher speeds, as an emergency stop would cause the target to quickly escape. In contrast, without strict safety constraints, Vis. Tracker is relatively more robust to the mapping noise, although the increased latency still limits its performance at higher speeds. However, its visibility performance is inferior to Elas. Tracker (which will be discussed later), resulting in a higher proportion of tracking failures attributed to target missed. The target may escape the FOV when performing agile maneuvers, particularly at low speeds where sharp turns occur more frequently. Conversely, the target's path is smoother due to kinematic constraints at higher speeds, leading to a noteworthy success rate at faster velocities.
By comparison, our approach consistently achieves high success rates from 3 to 7 m/s, where even the evader with privileged information would struggle to reach this maximum speed. On the one hand, the straightforward design significantly reduces latency and accumulated errors. As evidenced by recent research on high-speed flight \cite{science, Newton}, the processing efficiency is critical for autonomous quadrotors with the speed increasing. On the other hand, benefiting from the data-driven nature of deep learning and privileged learning strategy, our method demonstrates comparable noise robustness to the evader with ground-truth map, while relying only on onboard sensor observations. Moreover, the multimodal predictions enable comprehensive exploration of the solution space and prevent getting trapped in infeasible local minima. Therefore, our method demonstrates superior agile tracking and obstacle avoidance capabilities in high-speed flight. However, without target trajectory prediction, the failures of our method are mainly caused by target occlusion as shown in Fig. \ref{tracking_compare}C.

\begin{figure*}[t]\centering
        \includegraphics[width=\linewidth]{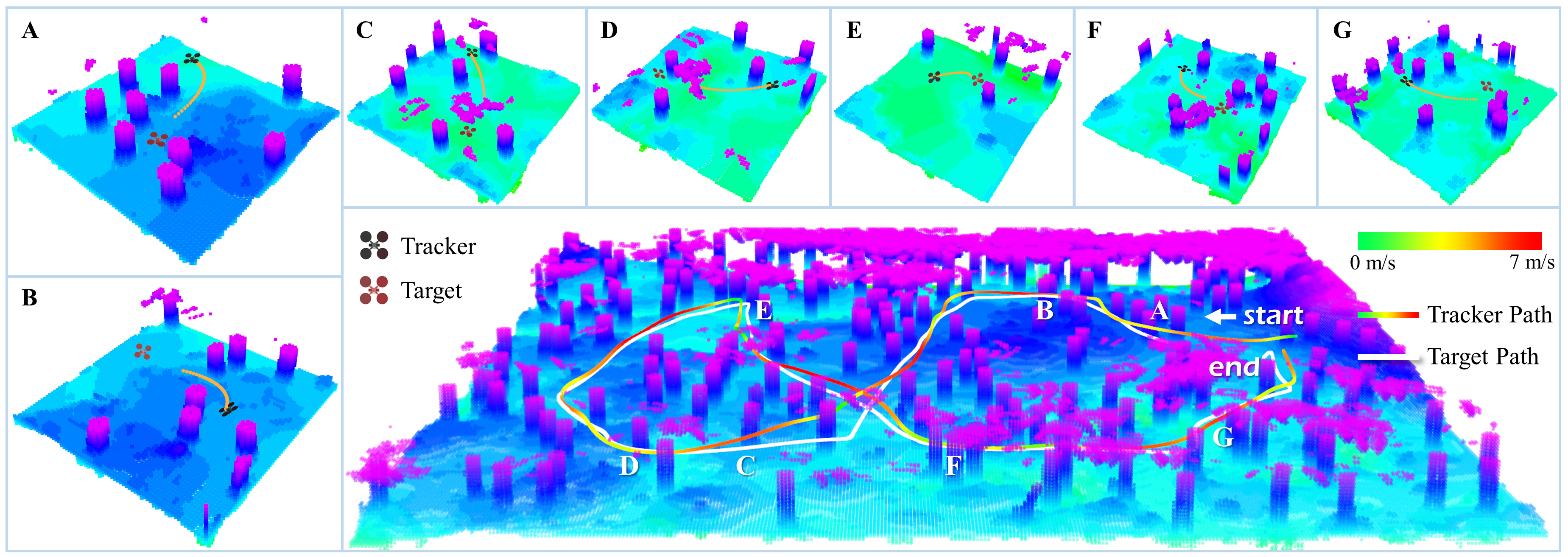}
        \caption{Visualization of large-scale tracking in simulation, where the target's path is drawn in white and the tracker's path is color-coded by speed. In snapshots A-G, the tracker's trajectory is represented in orange. Note that the map is only used for visualization and is not accessible to the tracker.}
        \label{largescale_tracking}
\end{figure*}

\begin{figure}[t]\centering
        \includegraphics[width=0.85\linewidth]{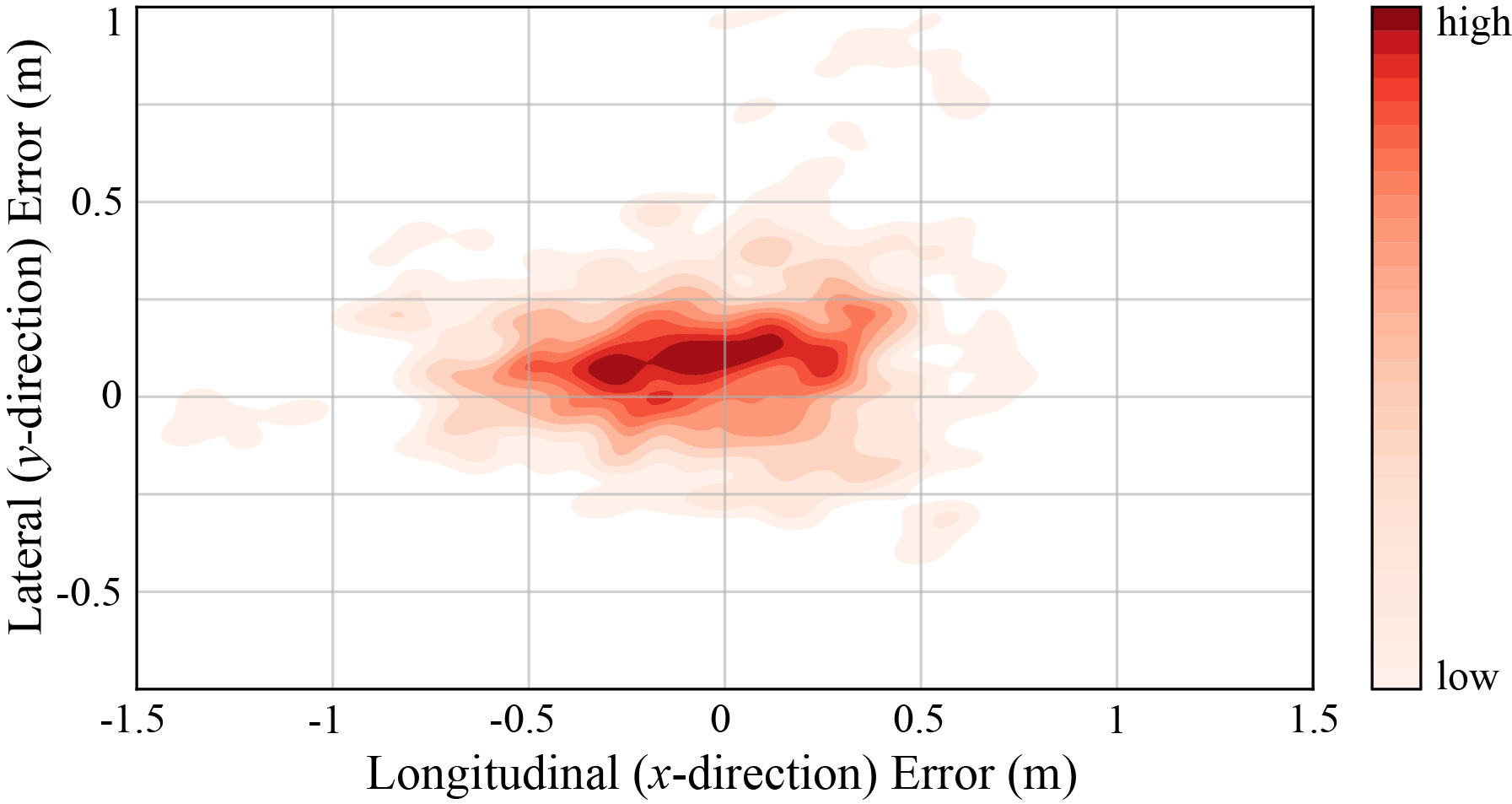}
        \caption{Distribution of target estimation error throughout tracking process.}\label{detection_error}
\end{figure}

We further visualize the target visibility of all methods during 3 m/s tracking. Specifically, we count the target positions projected to x-y plane in the tracking quadrotor's FOV, as shown in Fig. \ref{tracking_compare}D. The heat map shows the distribution of the target positions relative to the tracker and the red sector angle represents the FOV of the camera (set to 90° same as RealSense D455). Our results are consistent with those reported in \cite{elastic-tracker}, where the target distribution observed by Elas. Tracker is more concentrated than that of Vis. Tracker. This is because the Elas. Tracker imposes a hard constraint on the observation sectors, ensuring the target is stably observed at approximately 4 meters. However, due to the deceleration when reaching the limited distance and  braking caused by obstacles, the target has the chance to escape, leading to some outlier results. As a comparison, Vis. Tracker innovatively introduces a differentiable visibility cost by representing the FOV as a series of spheres. However, ensuring an obstacle-free FOV is sufficient but unnecessary for target visibility, especially in cluttered environments where the requirement is difficult to satisfy. Compared with the baselines, our method perceives with lower latency and directs the heading towards the estimated target without any intermediate steps. In addition, more efficient trajectory performance and greater clearance to obstacles can also help avoid occlusion to some extent, particularly facing rapid escape. Consequently, the target distribution of our method is more concentrated in the lateral direction. Due to the lack of explicit distance constraints to the target's future states, it is dispersed in the forward direction. Nevertheless, it remains within the 2-6 m safety range, as the fixed trajectory execution time enables automatic speed adjustments when the target is approached or moves away. Moreover, our visibility performance remains stable even at 7 m/s, despite the longer distance to target -- which is reasonable as higher speeds require a larger safety clearance.

We finally compare the distribution of smoothness cost (evaluated by the integral of squared jerk) and safety performance (evaluated by the minimum distance to obstacles) at 3 m/s over 10 trials, and additionally evaluate our method across different speeds. The results are illustrated in Fig. \ref{tracking_compare}E - F, respectively. The higher smoothness costs of baselines can be summarized as follows. The hard-constraint-based method \cite{elastic-tracker} involves braking and acceleration to maintain appropriate tracking distance or avoid sudden obstacles, as well as occasional winding trajectories in narrow corridors caused by noisy maps. To ensure obstacle-free FOV, \cite{visibility-tracker} incurs wide detours and sudden acceleration near obstacles, particularly in dense forests.
In comparison, our method achieves better smoothness and safety, and exhibits approximate performance at 7 m/s to that of the baselines at 3 m/s. There is only one instance of our method where the minimum clearance to obstacles is less than 0.2 m, which occurs at 7 m/s. As analyzed earlier, the proposed network policy, through privileged learning, can achieve comparable performance to privileged expert with only noisy onboard observations. Besides, we select the optimal prediction from a set of candidates considering both safety and smoothness costs, thus avoiding emergency braking and unnecessary deceleration. 

In conclusion, compared with traditional cascaded pipelines, our method offers substantial advantages in reducing both computational latency and error accumulation through straightforward design philosophy. Additionally, leveraging the data-driven nature to enhance the noise robustness and multimodal predictions for extensive exploration, it demonstrates superior capabilities in high-speed flight and agile tracking.

\subsubsection{Large-scale Tracking}

\begin{figure*}[t]\centering
        \includegraphics[width=\linewidth]{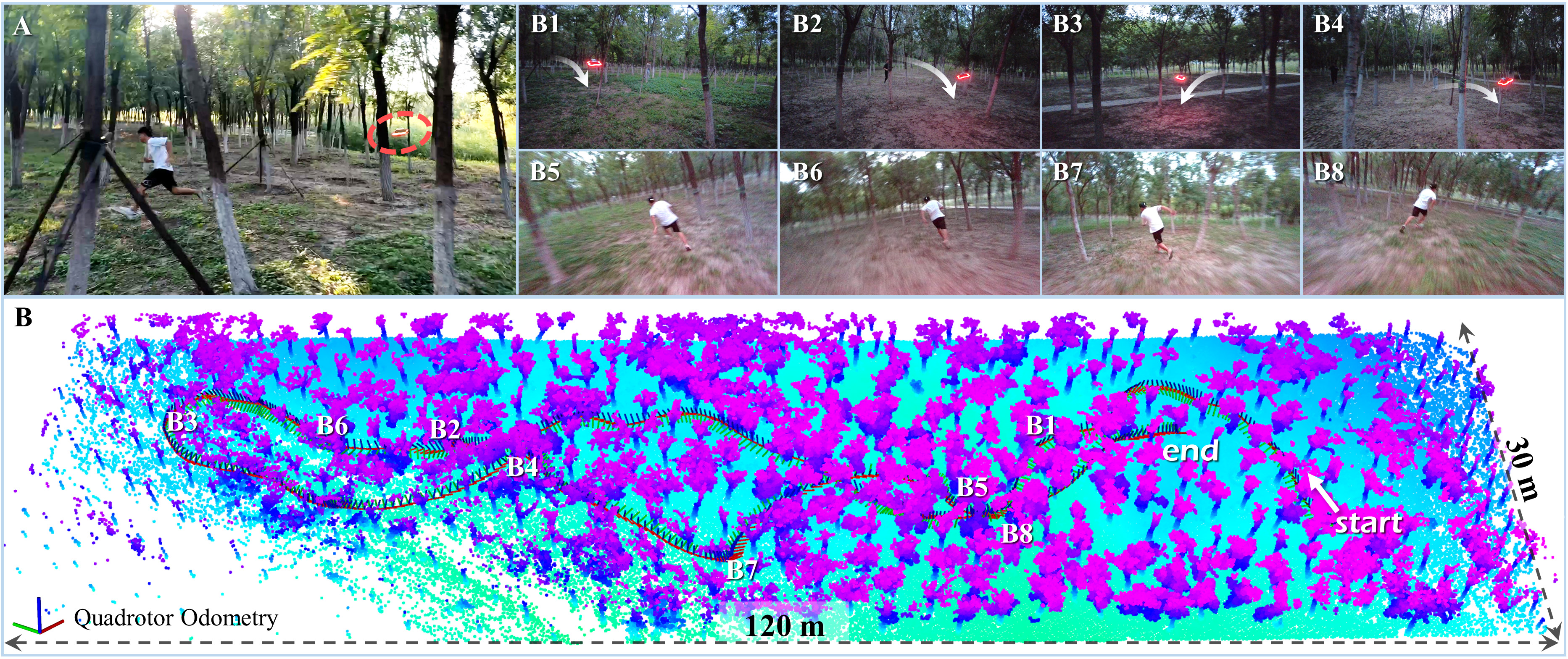}
        \caption{Agile tracking in real-world dense forests. (A) Experimental scenario with tree spacing of 3-4 m. (B) The tracking path of quadrotor in large-scale forest, where the maximum velocity reaches 6 m/s. (B1-B4) Snapshots of tracker captured by an action camera mounted behind the evader, with white arrows indicating the drone's maneuvers. (B5-B8) Snapshots of the evader recorded by onboard camera during tracking.}
        \label{forest_tracking}
\end{figure*}

We conducted a more challenging experiment in simulation to qualitatively demonstrate our performance, where the target fly along an $\infty$-shaped path in a large-scale map, with a maximum speed set to 7 m/s. Similarly, the ground-truth map is accessible to the target, while the tracker only has access to RGB-D images and its own odometry. The experimental scenario and complete paths of both evader and tracker are visualized in Fig. \ref{largescale_tracking}, where the tracker's path is color-coded by speed. The scenario includes sharp turns with large angles (e.g., Fig. \ref{largescale_tracking}E), occlusions and obstacles in dense areas (e.g., Fig. \ref{largescale_tracking}F), and rapid escapes in open areas, which pose considerable challenges to the tracker. As demonstrated by Fig. \ref{largescale_tracking}A-G, our method consistently maintains stable and aggressive tracking even facing challenges such as occlusions by obstacles, noisy depth observations, agile maneuvers and high-speed escapes of target.

Additionally, we plot the distribution of target state estimation errors throughout the entire tracking process in Fig. \ref{detection_error}, including the frames with occlusions. The error is calculated by projecting the target's positions of simulator's ground truth and our results to the x-y plane in tracker's FOV. For clarification, we use EKF to achieve smoother state estimation and compensate for the missed detections, without predicting the target's future trajectory. Besides, a larger process noise is used to prioritize the prediction of our network. As illustrated in Fig. \ref{detection_error}, the lateral (y-direction) error remains within ±0.5 m during 7 m/s tracking. The error in the longitudinal direction (i.e., the depth) is larger but still remains within the ±0.75 m range. This ensures the target stays at the center of FOV by directing the heading towards its position.

\subsubsection{Real-world Experiment}

\begin{figure*}[t]\centering
        \includegraphics[width=\linewidth]{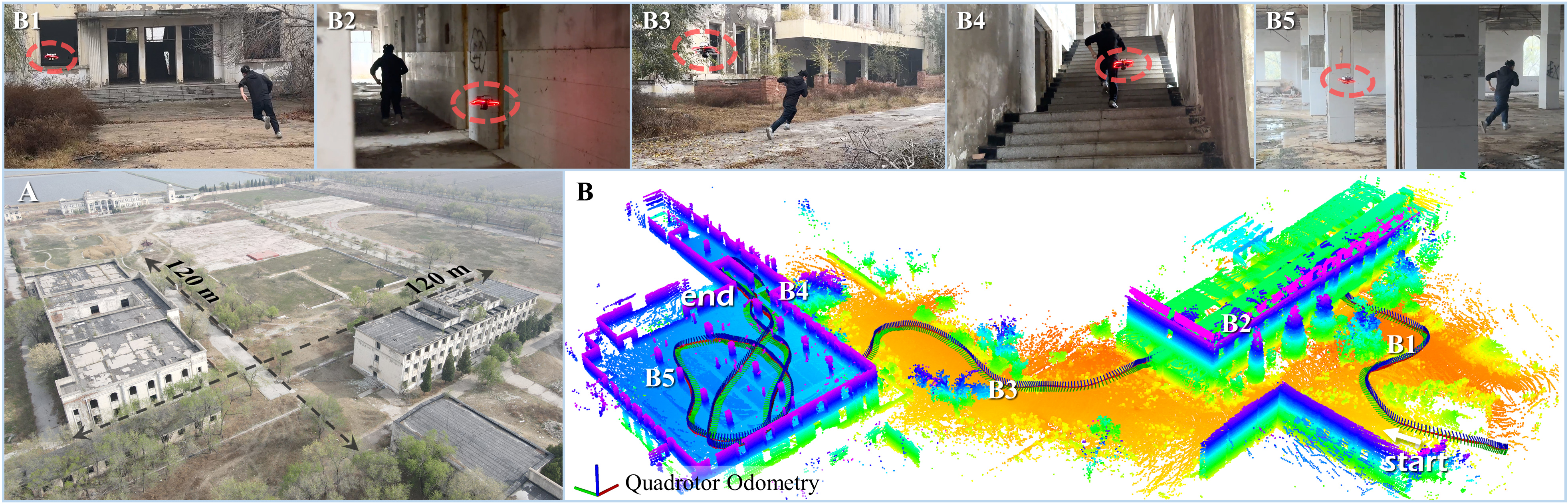}
        \caption{Agile tracking in real-world buildings. (A) Large-scale architectural scene spanning an area of 120$\times$120 m$^2$. (B) The tracking path of quadrotor represented by onboard visual odometry, where the maximum velocity reaches 5.6 m/s and the target's state as well as the map are completely inaccessible. (B1-B5) Snapshots of tracking process captured from a third-person perspective, corresponding to the marked positions in (B).}
        \label{building_tracking}
\end{figure*}

In this section, we first conduct target tracking experiments in a large-scale dense forest covering approximately 3600 m$^2$. As shown in Fig. \ref{forest_tracking}A, the environment is cluttered and unstructured, with trees spaced 3-4 m apart and dense low-hanging foliage throughout. In real-world experiments, we treat a human as the tracked target, which is highly agile and capable of swiftly weaving through the forest (e.g., Fig. \ref{forest_tracking}B5) and performing evasive maneuvers such as sudden turns (e.g., Fig. \ref{forest_tracking}B7). Different from previous works, the tracker operates without any prior information about the evader, who follows no predetermined escape strategy and has no interaction with the tracker. The only assumption is that the target is visible at the mission start, which can be seamlessly integrated with upstream exploration task \cite{fuel}. Moreover, the dense forest introduces unavoidable occlusions and poses substantial challenges to the safety of the quadrotor, as any braking may result in permanent loss of the target. Furthermore, larger-scale environment imposes greater demands on the algorithm's robustness and reliability, requiring the drone to closely follow the target constantly while swiftly avoiding unforeseen obstacles.

The results are demonstrated in Fig. \ref{forest_tracking}B and the supplementary video, where the tracker's path is obtained through visual odometry and visualized on the map with coordinate axes, while the target's position is unavailable. Moreover, the environment map is used solely for visualization, while only noisy RGB-D images are available to the quadrotor. The proposed system is fully onboard and achieves an average latency of only 8.2 ms from perception to action on NVIDIA Orin NX. The snapshots B1-B4 are captured by an action camera mounted behind the evader to illustrate the tracker from the evader's perspective. The white arrows indicate the agile maneuvers of the tracker in response to the target's sudden directional changes and evasive actions. The snapshots B5-B8 are the RGB observations of onboard camera corresponding to the marked positions in Fig. \ref{forest_tracking}. As visualized, the quadrotor achieves high-speed and long-range tracking despite the external challenges (such as cluttered environment, agile target, and low-light dusk) and internal limitations (such as motion blur, noisy depth images, and limited onboard computational resources). We achieve a maximum speed of 6 m/s, which is over 2 times faster than existing state-of-the-art methods in cluttered real-world scenarios (with a maximum of 2.5 m/s reported in \cite{visibility-tracker}). These results underscore the superior tracking capabilities of our method in complex unknown environments.

Subsequently, we conduct experiments in an abandoned school (as shown in Fig. \ref{building_tracking}A) to demonstrate the performance of our algorithm in complex architectural environments. The setup is consistent with the experiment in the wild, where the drone operates purely based on onboard computation and sensor, without access to any external or ground-based auxiliary information. As illustrated in Fig. \ref{building_tracking}B1-B5, the target first passes through bushes and dirt mounds, traverses the teaching building, then accelerates across an open area, goes upstairs to the second-floor cafeteria, and circles around inside. It encompasses more diverse environments, including narrow indoor corridors, open outdoor campus, and multi-level structures with staircases, posing a significant challenge for persistent tracking. The experimental results in Fig. \ref{building_tracking}B demonstrate that our method can reliably and persistently track an uncooperative target at high speed. In addition to the aforementioned difficulties in the jungle, the tracker further encounters narrow doorways and drastic lighting changes caused by indoor-outdoor transitions. In conclusion, a series of challenging real-world experiments verify the effectiveness and practicality of our method in high-speed and agile tracking across a variety of environments. 

\subsection{Navigation Tasks}
\subsubsection{Simulation Comparison}

\begin{figure*}[t]\centering
        \includegraphics[width=\linewidth]{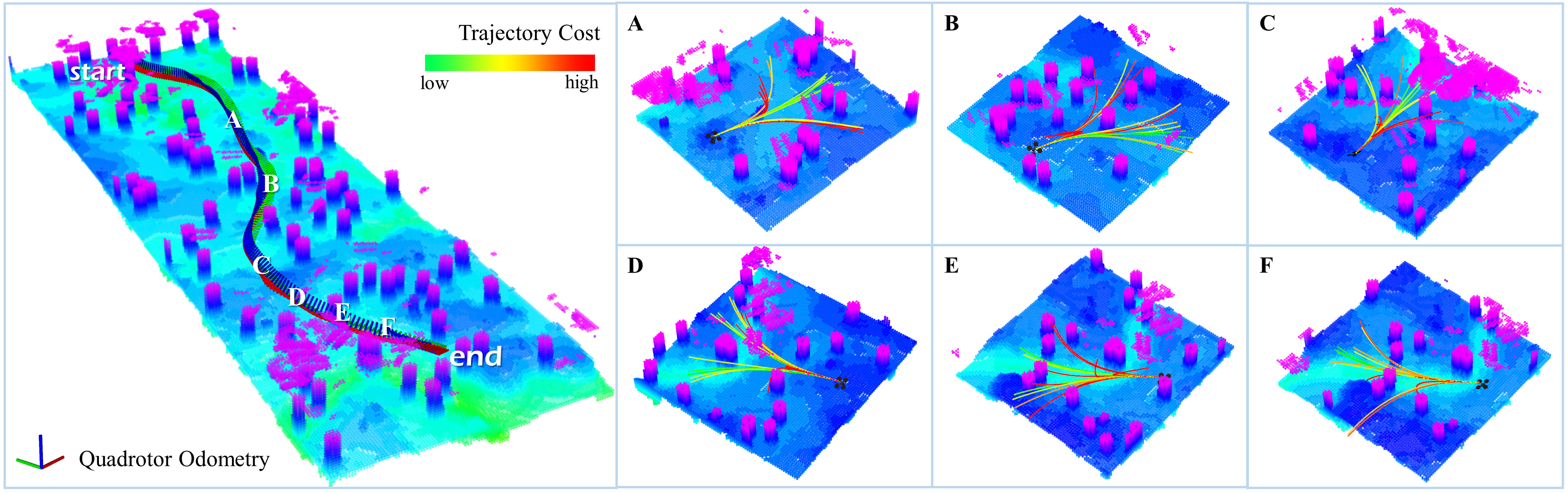}
        \caption{Visualization of high-speed navigation at 10 m/s in simulation, where the executed path is depicted by odometry axes. In snapshots A-F, all predicted trajectories are visualized and color-coded by corresponding costs. Note that the map is unavailable and only the optimal trajectory is solved in practice.}
        \label{sim_navigation}
\end{figure*}

\begin{figure}[t]\centering
        \centering
            \subfigure[Clearance of 4 m.]{\includegraphics[width=0.45\linewidth]{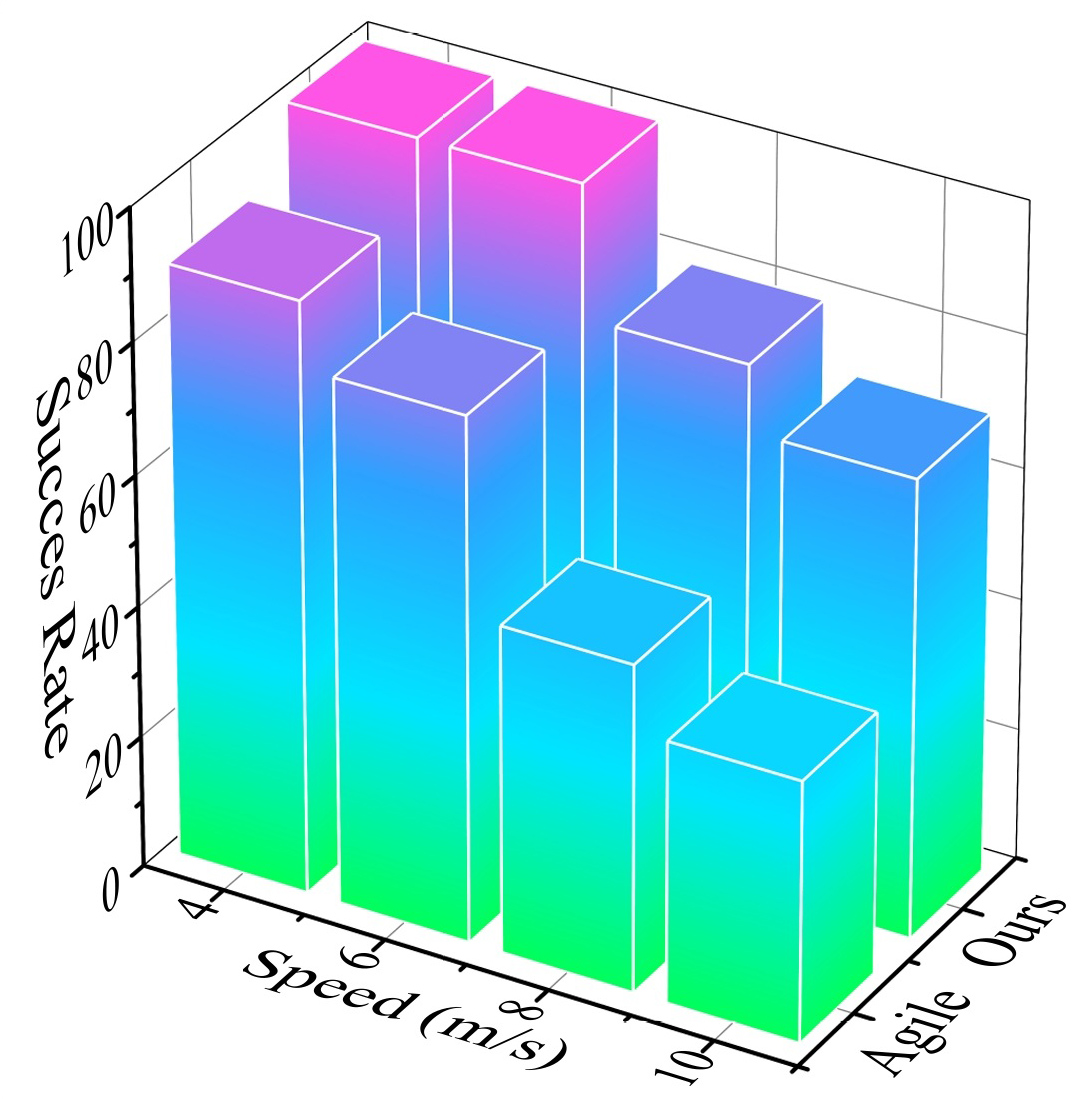}}
            \hspace{0.1cm}
            \subfigure[Clearance of 5 m.]{\includegraphics[width=0.45\linewidth]{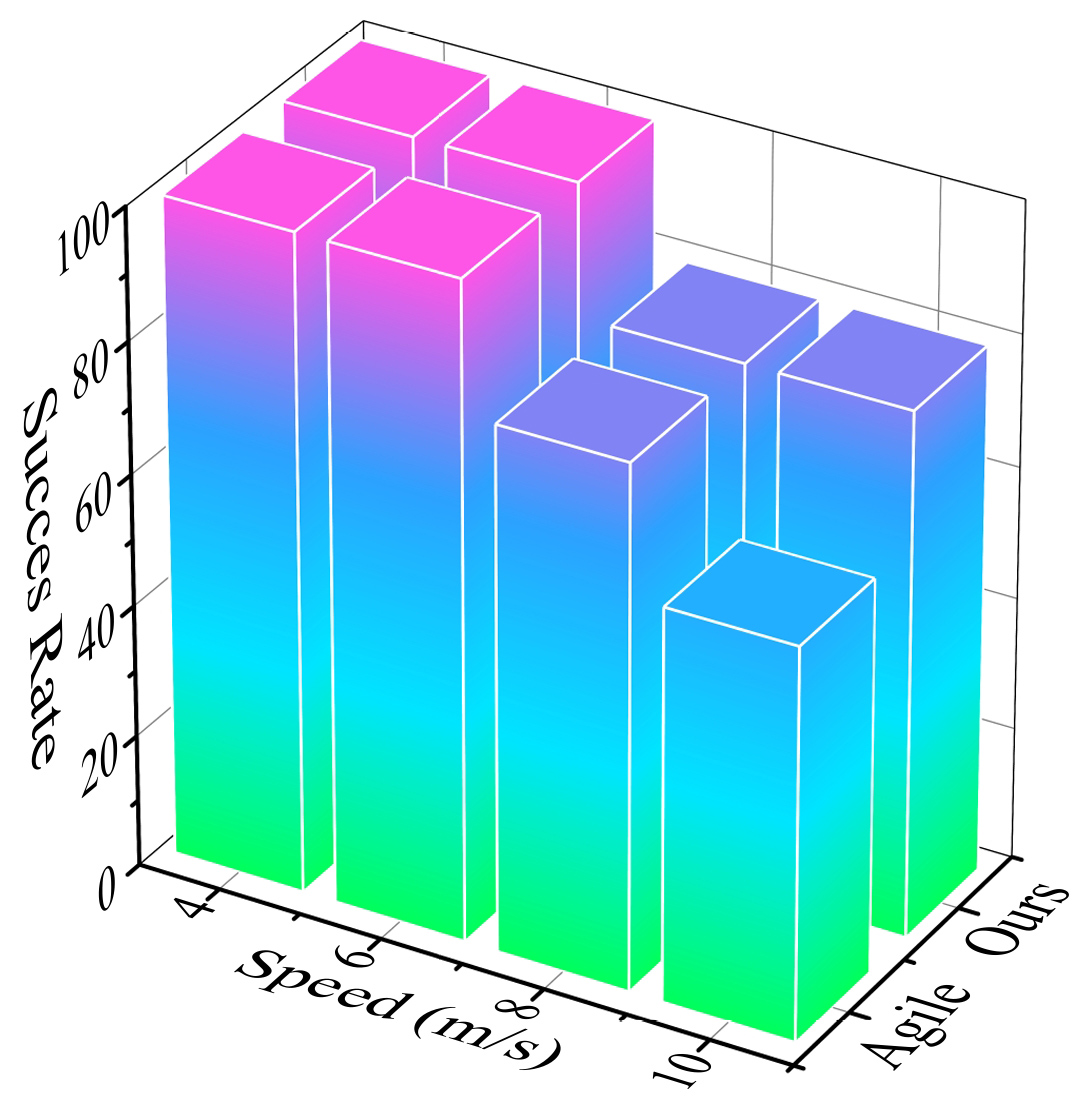}}
        \caption{Success rate comparison at various speeds and clearances.}
        \label{highspeed_compare}
\end{figure}

As illustrated in Fig. \ref{system_overview}, our framework can easily degenerate into a generic navigation policy by removing the target-related prediction and providing a desired flight direction as input. To further demonstrate the scalability of the proposed algorithm, additional experiments are conducted in dense environments to validate its high-speed navigation capability. We compare the success rate with the state-of-the-art open-source vision-based planner \cite{science} (called Agile), which has made groundbreaking advancements in high-speed flight. We conduct experiments in denser randomized forests, where the average spacing is 4 m and 5 m (corresponding to densities of 1/16 and 1/25 tree/m$^2$, respectively). The quadrotor's forward speeds vary between 4 m/s and 10 m/s, and we repeat the experiments 10 trials for each difficulty level with different realizations of the forest. The goal is set 40 m ahead and the trial is considered successful if the drone reaches the goal without collision. As reported in Fig. \ref{highspeed_compare}, our method is also competitive in high-speed navigation tasks, especially demonstrating superior performance compared to the baseline in obstacle-dense environments. 
The results can be interpreted as follows. Inspired by detection networks with thousands of predictions, our structure is inherently multimodal and can better explore the solution space through a set of anchor primitives, thereby providing more feasible candidates. Additionally, the anchor-based prediction ensures a clear correspondence between the trajectory and image features, preventing numerical instability and mode collapse (where all predictions converge to the same outcome). On the other hand, our expert-free training strategy seamlessly integrates traditional trajectory optimization with gradient descent in network training, which provides more realistic evaluation, eliminates label assignment, and supports a greater number of predictions. This expert-free strategy also simplifies data collection, facilitating extensive random sampling and data augmentation without the need for online rendering or re-annotation. Furthermore, the perception-to-action design enhances coordination between planning and control, eliminating the error accumulation of cascaded control structure.

We visualize a flight trial at 10 m/s with an average clearance of 4 m in Fig. \ref{sim_navigation} and illustrate the multimodal prediction in snapshots A-F. As shown by the odometry axes, the quadrotor performs aggressive maneuvers while navigating through dense obstacles. In snapshots Fig. \ref{sim_navigation}A-F, the predicted trajectories are color-coded by corresponding cost and nearly all of them are feasible. We select the optimal prediction based on the cost considering smoothness and safety performance to ensure temporal consistency, as topology switching will lead to large jerks and smoothness cost.

\subsubsection{Real-world Experiment}

\begin{figure*}[t]\centering
        \includegraphics[width=\linewidth]{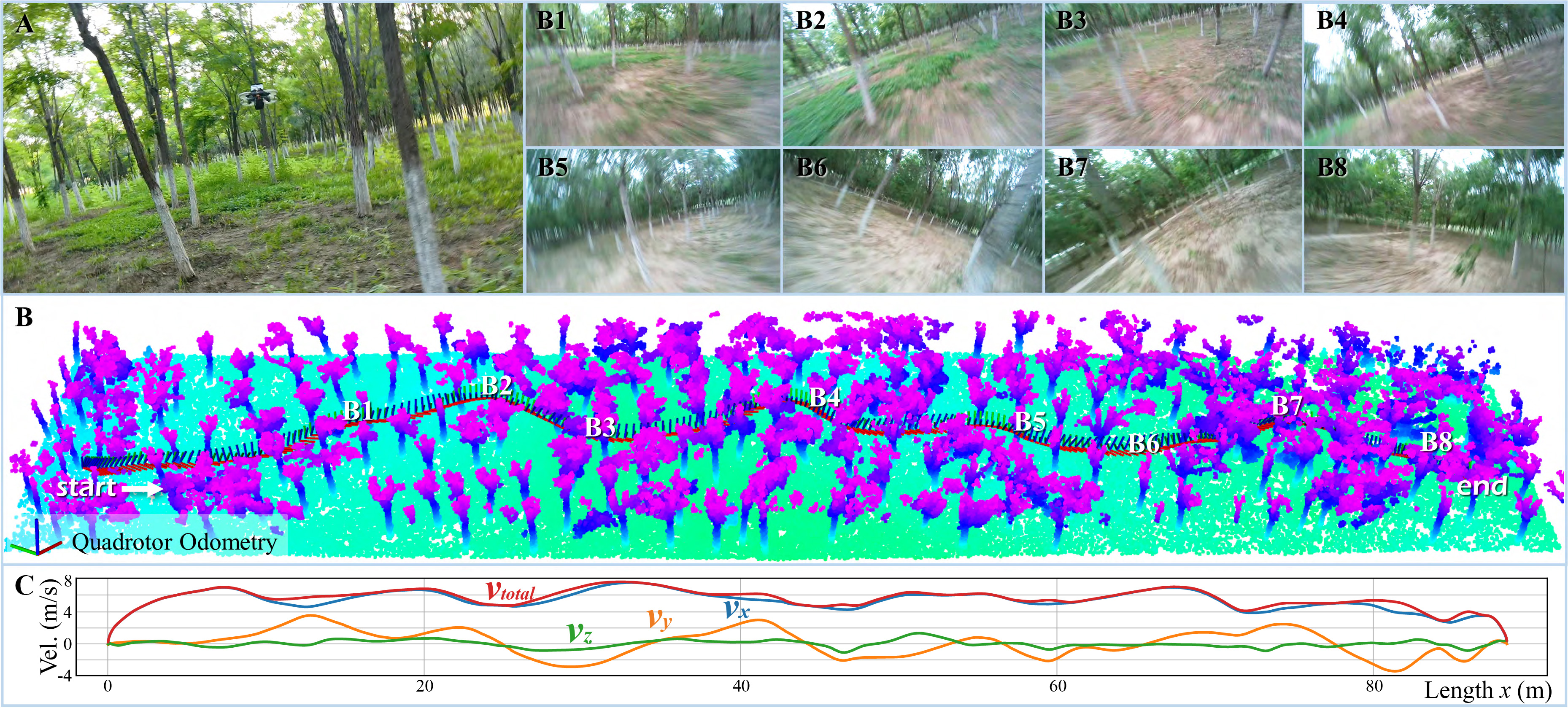}
        \caption{Real-world Experiment 1. (A) Experimental scenario with tree spacing of 3-4 m. (B) Visualization of flight path during high-speed navigation, with the goal set 90 m ahead. (B1-B8) Snapshots captured from the onboard camera during flight, corresponding to the marked positions in (B). (C) Speed profiles along each axis. The map is unknown to the quadrotor and only constructed for visualization.}
        \label{realworld_navigation_A}
\end{figure*}

\begin{figure}[t]\centering
        \includegraphics[width=\linewidth]{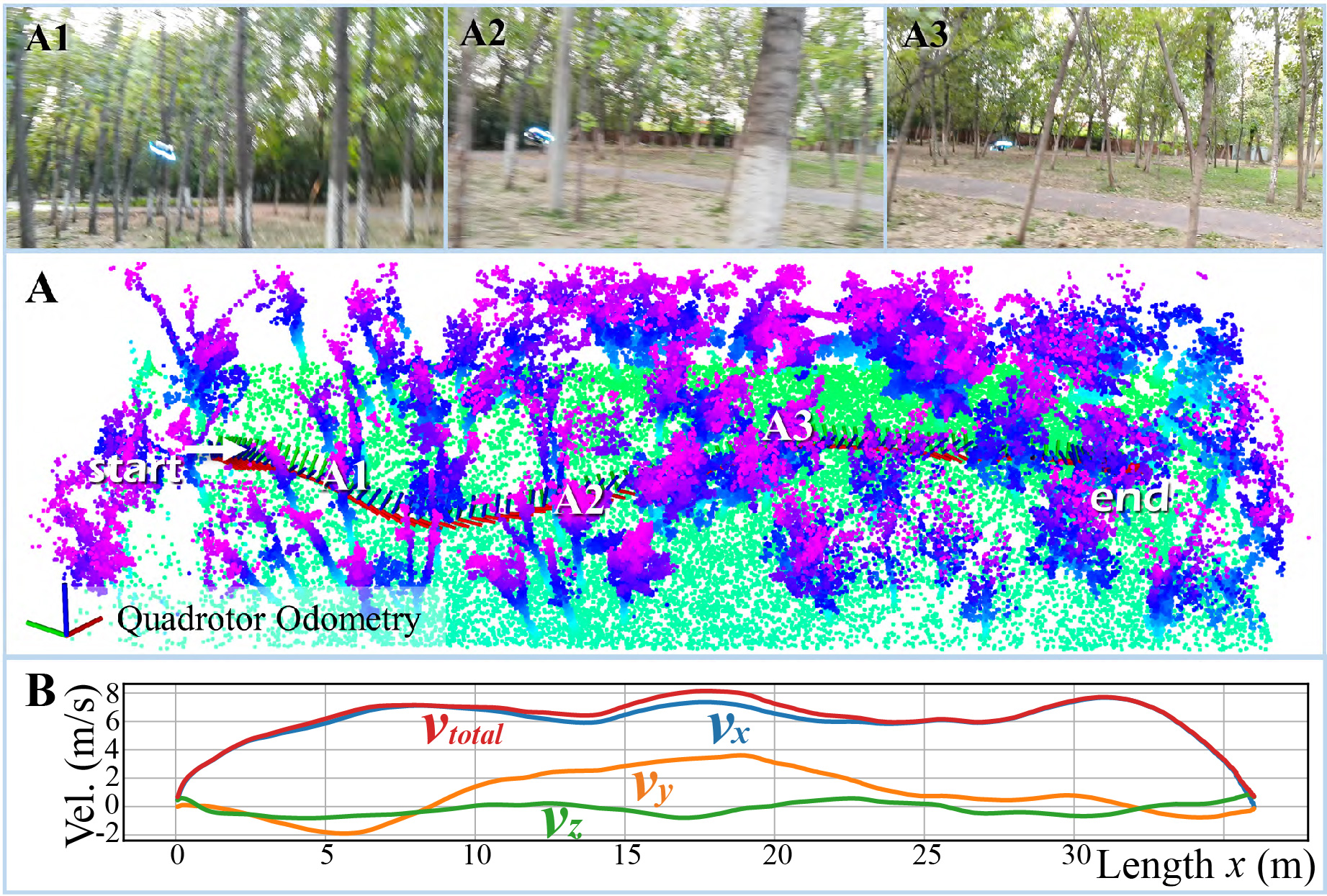}
        \caption{Real-world Experiment 2. (A) Visualization of flight path with the goal set 35 m ahead. (A1-A3) Snapshots captured from a third-person view during flight, corresponding to the marked positions in (A). (B) Speed profiles along each axis, with a maximum speed exceeding 8 m/s.}
        \label{realworld_navigation_B}
\end{figure}

In this section, we demonstrate the navigation performance of our policy in cluttered real-world environments. The previous methods \cite{science, Newton} have achieved remarkable maximum speeds in relatively sparse environments, significantly pushing the speed limits of vision-based planners. By contrast, we highlight the competitiveness of our multimodal approach in much denser forests through a series of high-speed flights.

We first conduct experiments in a large-scale wild forest, as shown in Fig. \ref{realworld_navigation_A}A. The environment is highly cluttered and obstacle-dense, with trees spaced approximately 3-4 m apart and low-hanging foliage. The goal is set 90 meters ahead, and the maximum flight speed is set to 8 m/s. The quadrotor operates completely autonomously during flight, relying solely on limited vision sensor and onboard computational resources. These pose significant challenges to the real-time performance and reliability of the navigation algorithm, as it must respond immediately to address the abrupt and unexpected obstacles. Fig. \ref{realworld_navigation_A}B demonstrates the flight path recorded by visual odometry, which is agile and aggressive. Snapshots Fig. \ref{realworld_navigation_A}B1-B8, captured by the onboard camera, correspond to the specific positions marked along the flight path. Additionally, the velocity profiles along each direction are shown in Fig. \ref{realworld_navigation_A}C, which are plotted in world coordinate system and take the forward length as horizontal axis to align with the Fig. \ref{realworld_navigation_A}B. 
We subsequently conduct flight tests in another smaller but denser forest, where the minimum safety clearance was less than 2 m (as shown in Fig. \ref{realworld_navigation_B}A1). The goal is set 35 meters ahead and the maximum flight speed is set to 8 m/s as well. The flight path is demonstrated in Fig. \ref{realworld_navigation_B}A, and the corresponding snapshots Fig. \ref{realworld_navigation_B}A1-A3 are captured from a third-person view during flight.  Complete demonstrations can be found in the supplementary video. In conclusion, a series of real-world experiments indicate that our method exhibits comparable performance to the state-of-the-art methods for high-speed navigation in denser environments. Besides, we avoid the conflict between actual perception and reference state for planning, reduce error accumulation inherent in cascaded structure, and support greater maneuverability by outputting thrust attitude as commands, thereby achieving a faster flight speed compared to our previous work \cite{yopo} (where the position error exceeds 0.5 m at 5.5 m/s flights).

\begin{figure}[t]\centering
        \includegraphics[width=\linewidth]{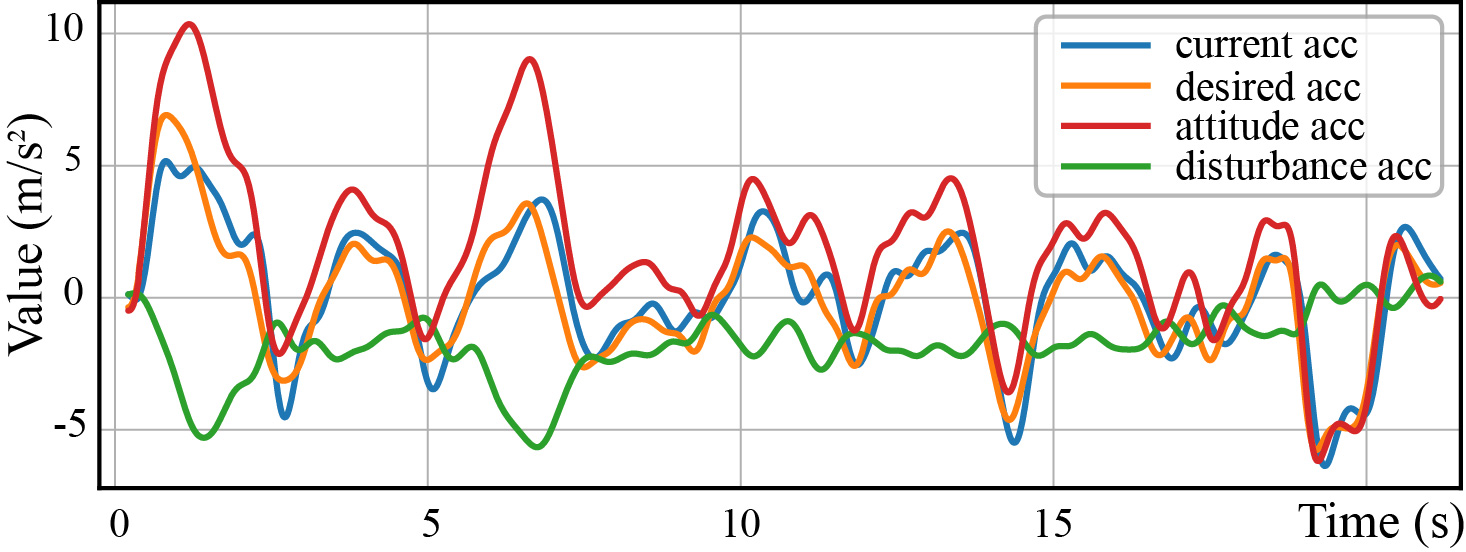}
        \caption{Measurements of the disturbance observer, desired acceleration, and actual acceleration in the x-direction during flight corresponding to Fig. \ref{realworld_navigation_A}.}
        \label{realworld_disturbance}
\end{figure}

Compared to predicting control commands by end-to-end network, we still plan the trajectory in the differentially flat space, which facilitates prediction evaluation and inherently ensures the smoothness of motion. Additionally, our method is independent of the quadrotor dynamics, thus eliminating the model mismatches between simulation and real-world deployment, enabling seamless transfer to different physical platforms. We use an idealized point-mass model during training without introducing any disturbances or model changes for domain randomization, nor any identification of actual physical platforms. In contrast, we employ a disturbance observer to online estimate the lumped disturbance primarily caused by aerodynamic drag force and external disturbances, relying solely on the current state observation which can be easily obtained from visual odometry. The effectiveness of the disturbance observer is visualized in Fig. \ref{realworld_disturbance}. Considering the dynamic of quadrotor in equation (\ref{dynamic}), we reformulate both the control commands and external disturbances into acceleration domain to enable consistent representation within the flatness-based framework. Specifically, in Fig. \ref{realworld_disturbance}, the desired acceleration $\ddot{\bm{f}}$ provided by planner is represented as desired acc, the ideal acceleration resulting from attitude control commands $({}^w\!\bm{R}_bF_c{\bm{e}_z} - \bm{g})/m$ is denoted as attitude acc, and $\bm{d}/m$ is the acceleration caused by lumped disturbance (written as disturbance acc). The attitude acceleration (the red curve), obtained by subtracting the disturbance acceleration (the green) from the desired acceleration (the orange), is used for control commands computation. The blue curve shows the current acceleration estimated by the onboard IMU. As visualized, taking the attitude acceleration as commands, the current acceleration can closely follow the desired values. It indicates that the system can effectively handle external disturbances and model uncertainties by compensating the disturbances into the network prediction for control commands calculation. In addition, the network performs inference in real-time based on the latest observations to cope with sudden obstacles in cluttered environments, providing feedback to ensure safety further. Moreover, it is worth noting that during the 8 m/s flight, the attitude accelerations mainly remain within 5 m/s$^2$, corresponding to a pitch angle below approximately 26 degrees. Therefore, we tilt the depth sensor by 20° to ensure that the camera would look forward during flight.

\section{Conclusion}
In this work, we propose an end-to-end agile tracking and navigation framework for quadrotors that directly maps the sensory observations to low-level control commands. Instead of making improvements to existing planning framework to make the pipeline more comprehensive and complex, we strive to simplify the pipeline and enable an instinctive, biologically inspired perception-to-action process for agile tracking. Our architecture is designed straightforward but interpretable, which explicitly incorporates detection, search, and optimization of the traditional pipeline into a single network. We address the mismatch between the physical model and simulation by disturbance observer and transform the predictions into commands to reduce the error accumulation of cascade control. We seamlessly bridge traditional trajectory optimization with deep learning by directly back-propagating the gradient of the trajectory cost to the network, enabling end-to-end training without expert supervision or online simulator interaction. Finally, a series of real-world tracking and high-speed navigation experiments are conducted to validate the efficiency of our algorithm.

\bibliographystyle{IEEEtran}
\bibliography{YOPOv2.bib}

\newpage

\vspace{11pt}

\vfill

\end{document}